\documentclass[letterpaper,11 pt]{article} 
\usepackage{amsmath,amsfonts,amssymb,color}
\usepackage{mathtools}
\usepackage{dsfont}
\usepackage[dvipsnames]{xcolor}
\usepackage{scalerel}
\definecolor{mygreen}{rgb}{0.0, 0.5, 0.0}
\definecolor{winered}{rgb}{0.8,0,0}
\definecolor{myblue}{rgb}{0,0,0.8}
\usepackage{tikz}
\usepackage{amsthm}
\usepackage{empheq}

\newtheorem{problem}{Problem}

\newtheorem{definition}{Definition}
\newtheorem{theorem}{Theorem}
\newtheorem{lemma}{Lemma}

\newtheorem{corollary}{Corollary}
\newtheorem{remark}{Remark}

\DeclarePairedDelimiter\ceil{\lceil}{\rceil}
\DeclarePairedDelimiter\floor{\lfloor}{\rfloor}

\DeclareMathOperator*{\argmax}{\arg\!\max}
\DeclareMathOperator*{\argmin}{\arg\!\min}

\newcommand{\mc}{\mathcal}
\newcommand{\mb}{\mathbf}
\DeclarePairedDelimiterX{\norm}[1]{\lVert}{\rVert}{#1}

\usepackage{algorithm}
\usepackage{algpseudocode}
\usepackage{epsfig}
\usepackage{array}
\usepackage{multirow}
\usepackage{epstopdf}
\usepackage{tikz}
\usepackage{relsize}
\usetikzlibrary{shapes,arrows}
\usepackage[hidelinks]{hyperref}
\hypersetup{
    colorlinks=true,
    linkcolor={winered},
    citecolor={myblue}
}
\usepackage{cite}
\usepackage[margin=1in]{geometry}
\title{\LARGE Exploiting Heterogeneity in Robust Federated Best-Arm Identification}
\author{Aritra Mitra,  George J. Pappas, and Hamed Hassani
\thanks{The authors are with the Department of Electrical and Systems Engineering, University of Pennsylvania. Email: {\tt \{amitra20, pappasg, hassani\}@seas.upenn.edu}. This work was supported by NSF Award 1837253, NSF CAREER award CIF 1943064, and the Air Force Office
of Scientific Research Young Investigator Program (AFOSR-YIP) under award FA9550-20-1-0111.}}
\date{}
\begin{document}
\maketitle
\thispagestyle{empty}
\pagestyle{empty}
\begin{abstract}
We study a federated variant of the best-arm identification problem in stochastic multi-armed bandits: a set of clients, each of whom can sample only a subset of the arms, collaborate via a server to identify the best arm (i.e., the arm with the highest mean reward) with prescribed confidence. For this problem, we propose \texttt{Fed-SEL}, a simple communication-efficient algorithm that builds on successive elimination techniques and involves local sampling steps at the clients. To study the performance of \texttt{Fed-SEL}, we introduce a notion of arm-heterogeneity  that captures the level of dissimilarity between distributions of arms corresponding to different clients. Interestingly, our analysis reveals the benefits of arm-heterogeneity in reducing both the sample- and communication-complexity of \texttt{Fed-SEL}. As a special case of our analysis, we show that for certain heterogeneous problem instances, \texttt{Fed-SEL} outputs the best-arm after just one round of communication. Our findings have the following key implication: unlike federated supervised learning where recent work has shown that statistical heterogeneity can lead to poor performance, one can provably reap the benefits of both local computation and heterogeneity for federated best-arm identification. As our final contribution, we develop variants of \texttt{Fed-SEL}, both for federated and peer-to-peer settings, that are robust to the presence of Byzantine clients, and hence suitable for deployment in harsh, adversarial environments. 
\end{abstract}

\section{Introduction}
Federated Learning (FL) has recently emerged as a popular framework for training statistical models using diverse data available at multiple clients (mobile phones, smart devices, health organizations, etc.) \cite{konevcny,mcmahan,bonawitz,surv1,surv2}. Two key challenges intrinsic to FL include statistical heterogeneity that arises due to differences in the data sets of the clients, and stringent communication constraints imposed by bandwidth considerations. Several recent works \cite{li,malinovsky,charles,charles2,fedsplit,scaffold} have reported that under statistical heterogeneity, FL algorithms that perform local computations to save communication can exhibit poor performance. Notably, these observations concern the task of supervised learning, which has almost exclusively been the focus of FL thus far.

In a departure from the standard supervised learning setting, the authors in \cite{dennis} recently showed that for the problem of federated clustering, one can in fact benefit from a notion of heterogeneity defined suitably for their setting. This is a rare finding that naturally begs the following question.

\vspace{1mm}
\textit{Are there other classes of cooperative learning problems where, unlike federated supervised learning, local computations and specific notions of heterogeneity can provably help?}
\vspace{1mm}

In this paper, we answer the above question in the affirmative by introducing and studying a federated variant of the \textit{best-arm identification} problem \cite{even2002,even,jamieson,audibert,bubeck,gabillon,jamieson2} in stochastic multi-armed bandits. In our model, a set of arms is partitioned into disjoint arm-sets and each arm-set is associated with a unique client. Specifically, each client can only sample arms from its respective arm-set. This modeling assumption is consistent with the standard FL setting where each client typically has access to only a portion of a large data set; we will provide further motivation for our model shortly. The task for the clients is to coordinate via a central server and identify the best arm with high probability, i.e., with a given fixed confidence.\footnote{We will provide a precise problem formulation in Sec. \ref{sec:problem} where we explain what is meant by a ``best arm".} To achieve this task, communication is clearly \textit{necessary}  since each client can only acquire information about its own arm-set in isolation. Our formulation naturally captures statistical heterogeneity: other than the requirement of sharing the same support, we allow distributions of arms across arm-sets to be arbitrarily different. In what follows, we briefly describe a couple of motivating applications.

\textbf{Applications in Robotics and Sensor Networks:} Our work is practically motivated by the diverse engineering applications (e.g., distributed sensing, estimation, localization, target-tracking, etc.) of cooperative learning \cite{loc1,franchi,savic,dames,dist_det}. For instance, consider a team of robots deployed over a large environment and tasked with detecting a chemical leakage based on readings from sensors spread out over the environment. Since it may be infeasible for a single robot to explore the entire environment, each robot is instead assigned the task of exploring a sub-region. The robots explore their respective sub-regions and coordinate via a central controller. We can abstract this application into our model by mapping robots to clients, controller to server, sensors to arms, and sub-regions to arm-sets. Based on this abstraction, detecting the sensor with the highest readings directly translates into the problem of identifying the best arm. 

\textbf{Applications in Recommendation Systems:} As another concrete application, consider a web-recommendation system for restaurants in a city. People in the city upload ratings of restaurants to this system based on their personal experiences. Based on these ratings, the system constantly updates its scores, and provides recommendations regarding the best restaurant(s) in the city. To capture this scenario via our model, one can interpret people as clients, the web-recommendation system as the server, and restaurants as arms. 
Since it is somewhat unreasonable to expect that a person visits every restaurant in the city, the rationale behind a client sampling only a subset of the arms makes sense in this context as well. 

We have thus justified the reason for considering heterogeneous action spaces (arm-sets) for the clients. In contrast, as we discuss in Section \ref{sec:rel_work}, prior related works make the arguably unrealistic assumption of a common action space for all clients, i.e., it is assumed that each client can sample every arm. As far as we are aware, this is the first work to study a federated/distributed variant of the heterogeneous multi-armed bandit (MAB) best-arm identification problem. The core question posed by this problem is the following: \textit{Although a client can sample only a subset of the arms itself, how can it still learn the best arm via collaboration?}

In this context, we will focus on two important themes relevant to each of the  applications we described above: \textit{communication-efficiency} and \textit{robustness}. To be more precise, we are interested in designing best-arm identification algorithms that (i) minimize communication between the clients and the server, and (ii) are robust to the presence of clients that act adversarially. Minimizing communication is motivated by low-bandwidth considerations in the context of sensor networks, and privacy concerns in the context of recommendation systems. Moreover, for the latter application, we would ideally like to avoid scenarios where a few deliberately biased reviews culminate in a bad recommendation. The need for robustness has a similar rationale in distributed robotics and sensor networks. Having motivated the setup, we now describe our main {contributions}. 

\subsection{Contributions}
The specific contributions of this paper are as follows.

$\bullet$ \textbf{Problem Formulation:} Our first contribution is to introduce the \textit{federated best-arm identification problem} in Section \ref{sec:problem}, and its robust variants in Sections \ref{sec:robust_variant} and \ref{sec:MAB_network}. As we explained earlier in the introduction, the problems we study have diverse practical applications that span collaborative learning in sensor networks, distributed robotics, and recommendation systems. 

In \cite{dennis}, the authors had conjectured that 
specific notions of heterogeneity - together with careful
analyses - may provide benefits for a plethora of problems in federated learning. By identifying federated best-arm identification as one such problem, our work is one of the earliest efforts to formally support this conjecture. 

$\bullet$ \textbf{Algorithm:}  In Section \ref{sec:Fed-SEL}, we propose \texttt{Fed-SEL} - a simple, communication-efficient algorithm that builds on successive elimination techniques \cite{even2002,even}, and comprises of two distinct phases. During Phase I, each client aims to identify the locally best arm in its own arm-set; this requires no communication. Phase II involves periodic, encoded  communication between the clients and the server (akin to standard FL algorithms) to determine the best arm among the locally best arms. As we discuss later in the paper, having two separate phases contributes towards (i) significantly reducing communication; and (ii) tolerating adversaries.

$\bullet$ \textbf{Benefits of Heterogeneity:} To analyze  \texttt{Fed-SEL}, we introduce a notion of \textit{arm-heterogeneity} in Section \ref{sec:results_FedSEL}. Roughly speaking, our definition of heterogeneity captures the ``closeness" between distributions of arms within an arm-set relative to that across different arm-sets. In Theorem \ref{thm:Fed-SEL}, we rigorously characterize the performance of \texttt{Fed-SEL}. Our analysis reveals that more the level of heterogeneity in a given instance of the problem, fewer the number of communication rounds it takes for \texttt{Fed-SEL} to output the best arm at the prescribed confidence level, i.e., \texttt{Fed-SEL} \textit{adapts} to the level of heterogeneity. We also identify problem instances where one round of communication suffices to learn the best arm. Since fewer rounds leads to fewer arm-pulls, we conclude that \textit{heterogeneity helps reduce both the sample- and communication-complexity} of \texttt{Fed-SEL}.

As far as we are aware, Theorem \ref{thm:Fed-SEL} is the only result to demonstrate provable benefits of heterogeneity in federated/distributed MAB problems. In the process of arriving at this result, we derive new bounds on the Lambert $W$ function \cite{lambertw} that may be of independent interest; see Lemma \ref{lemma:lambert} in Appendix \ref{app:lambert_bnds}.

$\bullet$ \textbf{Robustness to Byzantine adversaries:}  By now, there are several works that explore the challenging problem of tackling adversaries in supervised learning  \cite{chensu,blanchard,dong,draco,alistarhbyz,xie,lirsa,ghosh1,ghosh2,ghosh3}. However, there is no analogous result for the distributed MAB problem we consider here. We fill this gap by developing robust variants of \texttt{Fed-SEL} that output the best arm despite the presence of Byzantine adversarial clients, both in federated and peer-to-peer settings. Our robustness results, namely Theorems \ref{thm:rFedSEL} and \ref{thm:peertopeer}, highlight the role of information- and network structure-redundancy in combating adversaries. 

Overall, for the problem of cooperative best-arm identification, our work develops simple, intuitive algorithms that are communication-efficient, benefit from heterogeneity, and practically appealing since they can be deployed in harsh, adversarial environments. 

\subsection{Related Work}
\label{sec:rel_work}
In what follows, we position our work in the context of relevant literature. 

$\bullet$ \textbf{Multi-Armed Bandits in Centralized and Distributed Settings:} Best-arm identification is a classical problem in the MAB literature and is typically studied with two distinct objectives: (i) maximizing the probability of identifying the best arm given a fixed sample budget \cite{audibert,bubeck,gabillon}; and (ii) minimizing the number of arm-pulls (samples) required to identify the best arm at a given fixed confidence level \cite{even2002,even,jamieson}. In this paper, we focus on the fixed confidence setting; for an excellent survey on this topic, see \cite{jamieson2}.
 
 The best-arm identification problem has also been explored in distributed \textit{homogeneous} environments \cite{dpexp1,dexp2} where every agent (client) can sample all the arms, i.e., each agent can identify the best-arm \textit{on its own}, and communication serves to reduce the sample-complexity (number of arm-pulls) at each agent. A fundamental difference of our work with  \cite{dpexp1,dexp2} is that communication is \textit{necessary} for the heterogeneous setting we consider. Moreover, unlike the vast majority of distributed MAB formulations that focus on minimizing the cumulative regret metric \cite{dMAB1,kalathil,kar,landgren1,landgren2,shahrampour,dMAB2,kolla,dMAB3,sankararaman,martinez,dubeyICML20,dubeyNIPS20,lalitha20,chawla1,chawla2,ghoshbandits,agarwal21,FedMAB1,FedMAB3}, the goal of our study is best-arm identification within a PAC framework. 

A couple of recent works demonstrate that in the context of multi-agent stochastic linear bandits, if the parameters of the agents share a common low-dimensional representation \cite{chawla2}, or exhibit some other form of similarity \cite{ghoshbandits}, then such structure can be exploited to improve performance (as measured by regret). Thus, these works essentially show that \textit{similarity helps.} Our findings complement these results by showing that depending upon the problem formulation and the performance measure, \textit{dissimilarity can also help}. 

$\bullet$  \textbf{Robust Multi-Armed Bandits:} A recent stream of works has focused on understanding the impacts of adversarial corruptions in MAB problems, both in the case of independent arms \cite{junadv,liuadv,lykouris,guptaadv}, and also for the linear bandit setting \cite{bogunovic1,garcelon,bogunovic2}. In these works, it is assumed that an attacker with a bounded attack budget can perturb the stochastic rewards associated with the arms, i.e., the attack is \textit{structured} to a certain extent. The Byzantine attack model we consider in this work is different from the reward-corruption models in \cite{junadv,liuadv,lykouris,guptaadv,bogunovic1,garcelon,bogunovic2} in that it allows a malicious client to act \textit{arbitrarily}. Thus, the techniques used in these prior works do not apply to our setting. 

Like our work, worst-case adversarial behavior is also considered in \cite{vial}, where a \textit{blocking} approach is used to provide regret guarantees that preserve the benefits of collaboration when the number of adversaries is small. While the authors in \cite{vial} assume a complete graph and focus on regret minimization, our analysis in Section \ref{sec:MAB_network} accounts for general networks, and our goal is best-arm identification, i.e., we focus on the pure exploration setting.  

 $\bullet$ \textbf{Federated Optimization:} The standard FL setup involves periodic coordination between the server and clients to solve a distributed optimization problem \cite{konevcny,mcmahan,bonawitz,surv1}. The most popular algorithm for this problem, \texttt{FedAvg} \cite{mcmahan},  has been extensively analyzed  both in the homogeneous setting \cite{stich,wang,spiridonoff,FedPAQ,haddadpourNIPS, haddadpour2021federated,woodworth1}, and also under statistical heterogeneity \cite{khaled1,khaled2,haddadpour,li,koloskova}. Several  more recent works have focused on improving the convergence guarantees of \texttt{FedAvg} by drawing on a variety of ideas: proximal methods in \texttt{FedProx} \cite{sahu}; normalized aggregation in  \texttt{FedNova} \cite{gauri}; operator-splitting in \texttt{FedSplit} \cite{fedsplit}; variance-reduction in \texttt{Scaffold} \cite{scaffold} and \texttt{S-Local-SVRG} \cite{gorbunov}; and gradient-tracking in  \texttt{FedLin} \cite{FedLin}. Despite these promising efforts, there is little to no theoretical evidence that justifies performing local computations - a key feature of FL algorithms - under  statistical heterogeneity. 
 
 \textit{Heterogeneity can hurt:} Indeed, even in simple, deterministic settings with full client participation, it is now known that \texttt{FedAvg} and \texttt{FedProx} fail to match the basic centralized guarantee of linear convergence to the global minimum for strongly-convex, smooth loss functions \cite{li,malinovsky,charles,charles2,fedsplit}. Moreover, for algorithms such as \texttt{Scaffold} \cite{scaffold}, \texttt{S-Local-SVRG} \cite{gorbunov}, and \texttt{FedLin} \cite{FedLin} that do guarantee linear convergence to the true minimizer, their linear convergence rates demonstrate no apparent benefit of performing multiple local gradient steps under heterogeneity. To further exemplify this point, a lower bound is presented in \cite{FedLin} to complement the upper bounds in \cite{scaffold}, \cite{gorbunov}, and \cite{FedLin}; see also Remark 2.2 in \cite{gorbunov}. Thus, under heterogeneity, the overall picture is quite grim in FL. 
 
 In Section \ref{sec:problem}, we draw various interesting parallels between our formulation and the standard federated optimization setup. These parallels are particularly insightful since they provide intuition regarding why both local computations and heterogeneity can be helpful (if exploited appropriately) for the problem we study, thereby further motivating our work. 
\newpage
\section{Problem Formulation}
\label{sec:problem}
 Our model is comprised of a set of arms $\mathcal{A}$, a set of clients $\mathcal{C}$, and  a central server. The set $\mathcal{A}$ is partitioned into $N$ disjoint arm-sets, one corresponding to each client. In particular, client $i\in\mathcal{C}$ can only sample arms from the $i$-th arm set, denoted by $\mathcal{A}_i=\{{a^{(i)}_1},{a^{(i)}_2},\ldots,{a^{(i)}_{|\mathcal{A}_i|}}\}$. Throughout, we will make the mild assumption that each arm-set contains at least two arms, i.e., $|\mathcal{A}_i| \geq 2, \forall i\in[N]$. We associate a notion of time with our model: in each unit of time, a client can only sample one arm. Each time client $i$ samples an arm $\ell \in\mathcal{A}_i$, it observes a random reward $X_{\ell}$ drawn independently of the past (actions and observations of all clients) from a distribution $D_{\ell}$ (unknown to client $i$) associated with arm $\ell$. We will use 
$\hat{r}_{i,\ell}(t)$ to denote the average of the first $t$ independent samples $X_{i,\ell}(s), s\in [t]$ of the reward of an arm $\ell\in\mc{A}_i$ as observed by client $i$,  i.e., $\hat{r}_{i,\ell}(t) = (\sum_{s=1}^{t} X_{i,\ell}(s))/t$.

As is quite standard \cite{even2002,even,audibert,bubeck,gabillon,jamieson,jamieson2}, we assume that all rewards are in $[0,1]$. For each arm $\ell\in\mathcal{A}$, let $r_{\ell}$ denote the expected reward of arm $\ell$, i.e., $r_{\ell}=\mathbb{E}[X_{\ell}]$. We will use $\Delta_{\ell j} \triangleq r_{\ell}-r_{j}$ to represent the difference between the mean rewards of any two arms $\ell, j \in \mc{A}$. An arm with the highest expected reward in $\mathcal{A}_i$ (resp., in $\mathcal{A}$) will be called a \textit{locally best arm} in arm-set $i$ (resp., a \textit{globally best arm} in $\mathcal{A}$), and be denoted by $a^*_i$ (resp., by $a^*$). For simplicity of exposition, let the arms in each arm-set $\mathcal{A}_i$ be ordered such that $r_{a^{(i)}_1} > r_{a^{(i)}_2} > \cdots > r_{{a^{(i)}_{|\mathcal{A}_i|}}}$; thus, $a^*_i = a^{(i)}_1$.  Without loss of generality, we assume that arm $1$ is the \textit{unique} globally best arm, and that it belongs to arm-set $\mathcal{A}_1$, i.e., $a^*=a^{(1)}_1=1$. We can now concretely state the problem of interest.
 
 \begin{problem} \label{prob:FedMAB}
 (\textbf{Federated Best-Arm Identification}) Given any $\delta \in (0,1)$, design an algorithm that with probability at least $1-\delta$ terminates in finite-time, and outputs the globally best-arm $a^*$, where 
 \begin{equation}
  a^*=\argmax_{i\in\{1, \ldots, N\}, \ell\in\{1,\ldots,|\mc{A}_i|\}} \left(r_{a^{(i)}_{\ell}}=\mathbb{E}[X_{a^{(i)}_{\ell}}] \right). 
 \end{equation}
 \end{problem}

In words, given a fixed confidence level $\delta$ as input, our goal is to design an algorithm that enables the clients and the server to identify the globally best arm with probability at least $1-\delta$. Following \cite{even}, we will call such an algorithm $(0,\delta)$-PAC. Since each client $i\in\mathcal{C}$ is only \textit{partially informative} (i.e., can only sample arms from its local arm-set $\mathcal{A}_i$), it will not be able to identify $a^*$ on its own, in general. Thus, any algorithm that solves Problem \ref{prob:FedMAB} will necessarily require communication between the clients and the server. Given that the communication-bottleneck is a major concern in FL, our focus will be to design \textit{communication-efficient best-arm identification algorithms}. Later, in Section \ref{sec:robust_variant}, we will consider a \textit{robust} version of Problem \ref{prob:FedMAB} where the goal will be to identify the globally best arm despite the presence of adversarial clients that can act \textit{arbitrarily}. We now draw some interesting parallels between the above formulation and the familiar federated optimization setup that is used to model and study  supervised learning problems. 

\textbf{Comparison with the standard FL setting:} The canonical FL setting involves solving the following optimization problem via periodic communication between the clients and the server \cite{konevcny,mcmahan,bonawitz,surv1}:
\begin{equation}
  \min_{x\in\mathbb{R}^d}   f(x), \hspace{2mm} \textrm{where} \hspace{1mm} f(x) =\frac{1}{N}\sum_{i=1}^{N} f_i(x), \hspace{1mm} \textrm{and} \hspace{1mm}  f_i(x) = \mathbb{E}_{z_i \sim  {P}_i}[L_i(x;z_i)].
\label{eqn:objective}
\end{equation}
Here, $f_i:\mathbb{R}^{d}\rightarrow \mathbb{R}$ is the local loss function of client $i$, and $f(x)$ is the global objective function. Let $x^*_i=\argmin_{x\in\mathbb{R}^d} f_i(x)$, and $x^*=\argmin_{x\in\mathbb{R}^d} f(x)$. Under \textit{statistical heterogeneity}, $P_i \neq P_j$ in general for distinct clients $i$ and $j$, and hence, $f_i(x)$ and $f_j(x)$ may have different minima. A typical FL algorithm that solves for $x^*$ operates in rounds. Between two successive communication rounds, each client performs local training on its own data using stochastic gradient descent  (SGD) or a variant thereof. As a consequence of such local training, popular FL algorithms such as \texttt{FedAvg} \cite{mcmahan} and \texttt{FedProx} \cite{sahu}  exhibit a ``client-drift phenomenon": the iterates of each client $i$ drift-off towards the minimizer $x^*_i$ of its own local loss function $f_i(x)$  \cite{li,malinovsky,charles,charles2,fedsplit,scaffold}. Under statistical heterogeneity, there is no apparent relationship between $x^*_i$ and $x^*$, and these points may potentially be far-off from each other. Thus, client-drift can cause \texttt{FedAvg} and \texttt{FedProx} to exhibit poor convergence.\footnote{By poor convergence, we mean that these algorithms either converge slowly to the true global minimum, or converge to an incorrect point (potentially fast). See \cite{charles}, \cite{charles2}, and \cite{FedLin} for a detailed discussion on this topic.} More recent algorithms such as \texttt{SCAFFOLD} \cite{scaffold} and \texttt{FedLin} \cite{FedLin}  try to mitigate the client-drift phenomenon by employing techniques such as variance-reduction and gradient-tracking. However, even these more refined algorithms are unable to demonstrate any quantifiable benefits of performing multiple local steps under arbitrary statistical heterogeneity.\footnote{Indeed, in \cite{FedLin}, the authors show that there exist simple instances involving two clients with quadratic loss functions, for which,  performing multiple local steps yields no improvement in the convergence rate even when gradient-tracking is employed.}  

\begin{algorithm}[t]
\caption{Local Successive Elimination at Client $i$}
\label{algo:algo1}  
 \begin{algorithmic}[1]
\State \textbf{Parameters:} $\{\alpha_i(t),\beta_i(t)\}$, where 
$$ \alpha_i(t)=\sqrt{{\log\left({4|\mathcal{C}||\mc{A}_i|t^2}/{\delta}\right)}/{t}}; \beta_i(t) = c \alpha_i(t)$$
\State \textbf{Set:} $t=1$, $\mc{S}_i(t) = \mc{A}_i$
\State Sample each arm $\ell\in\mc{S}_i(t)$ once and  update empirical mean $\hat{r}_{i,\ell}(t)$ of arm $\ell$ \label{line:2}
\State Let $\hat{r}_{i,\max}(t) = \underset{\ell\in\mc{S}_i(t)}{\max} \  \hat{r}_{i,\ell}(t)$
\State Eliminate sub-optimal arms: $\mc{S}_i(t+1) = \{\ell\in\mc{S}_i(t) \ : \ \hat{r}_{i,\max}(t) - \hat{r}_{i,\ell}(t) < \beta_i(t)\}$
\State If {$|\mc{S}_i(t+1)|>1$}, then set $t=t+1$ and Go To Line \ref{line:2}. Else, \textbf{Output}  $a_i=S_i(t+1)$
\end{algorithmic}
 \end{algorithm}
 
Coming back to the MAB problem, $a^*_i$ can be thought of as the direct analogue of $x^*_i$ in the standard FL setting. We make two key observations.

\begin{itemize}
    \item \textbf{Observation 1:} Unlike the standard FL setting, there is an immediate relationship between the global optimum, namely the globally best arm $a^*$, and the locally best arms $\{a^*_i\}_{i\in\mc{C}}$: \textit{the globally best arm is the arm with the highest expected reward among the locally best arms.}
    
    \item \textbf{Observation 2:} Suppose distributions $D_{\ell_1}$ and $D_{\ell_2}$ corresponding to arms $\ell_1\in\mc{A}_i$ and $\ell_2 \in \mc{A}_j$ are ``very different" (in an appropriate sense). Intuition dictates that this should make the task of identifying the arm with the higher mean easier, i.e., \textit{statistical heterogeneity among distributions of arms corresponding to different clients should facilitate best-arm identification.} 
\end{itemize}

The first observation, although simple and rather obvious, has the following important implication. Suppose each client $i$ does pure exploration locally to identify the best arm $a^*_i$ in its arm-set $\mathcal{A}_i$. Clearly, such local computations contribute towards the overall task of identifying the globally best arm (as they help eliminate sub-optimal arms), regardless of the level of heterogeneity in the distributions of arms across arm-sets. \textit{Thus, we immediately note a major distinction between the standard federated supervised learning setting and the federated best-arm identification problem: for the former, local steps can potentially hurt under statistical heterogeneity; for the latter, we can exploit problem structure to  design local computations that always contribute towards the global task.} In fact, aligning with the second observation, we will show later in Section \ref{sec:results_FedSEL} that statistical heterogeneity of arm distributions across clients can assist in minimizing communication. Building on the insights from this section, we now proceed to develop a federated best-arm identification algorithm that solves Problem \ref{prob:FedMAB}.  

\newpage
\section{Federated Successive Elimination}
\label{sec:Fed-SEL}
In this section, we develop an algorithm for solving Problem \ref{prob:FedMAB}. Since minimizing communication is one of our primary concerns, let us start by asking what a client can achieve \textit{on its own} without interacting with the server. To this end,  consider the following example to help build intuition.

\textbf{Example:} Suppose there are just 2 clients with associated arm-sets $\mathcal{A}_1=\{1,2\}$ and $\mathcal{A}_2=\{3,4\}$. The mean rewards are $r_1=0.9, r_2=0.3, r_3=0.8$, and $r_4=0.7$. From classical MAB theory, we know that to identify the best arm among two arms with gap in mean rewards $\Delta$, each of the arms needs to be sampled $O(1/\Delta^2)$ times. Thus, after sampling each of the arms in $\mathcal{A}_1$ for $O(1/\Delta^2_{12})$ number of times (resp., arms in $\mathcal{A}_2$ for $O(1/\Delta^2_{34})$ times), client 1 (resp., client 2) will be able to identify arm 1 (resp., arm 3) as the locally best-arm in $\mathcal{A}_1$ (resp., $\mathcal{A}_2$) with high probability. However, at this stage, arm 1 will likely have been sampled much fewer times than arm 3 (since $1/\Delta^2_{12} \approx 3$, and $1/\Delta^2_{34} = 100$), and hence, client 1's estimate of arm 1's true mean will be much coarser than client 2's estimate of arm 3's true mean. The main message here is that \textit{simply identifying locally best arms from each arm-set and then naively comparing between them may result in rejection of the globally best-arm.} Thus, we need a more informed strategy that accounts for imbalances in sampling across arm-sets. With this in mind, we develop \texttt{Fed-SEL}. 

\begin{algorithm}[H]
\caption{Federated Successive Elimination (\texttt{Fed-SEL})}
\label{algo:algo2}  
 \begin{algorithmic}[1] 
\Statex \hspace{-5mm} \textbf{Phase I:} \textit{Identifying locally best arms}
\State Each client $i\in\mathcal{C}$ runs Algo. \ref{algo:algo1} on $\mathcal{A}_i$. If Algo. \ref{algo:algo1} terminates at client $i$, then client $i$ reports $\{a_i,\hat{r}_{i,a_i}(\bar{t}_i),\bar{t}_i\}$ to the server
\Statex \hrulefill
\Statex \hspace{-5mm} \textbf{Phase II:} \textit{Identifying the globally best arm}
 \For {$k\in\{0,1,\ldots\}$}
\Statex \hspace{-3mm} $\bullet$ Server initializes $\bar{\Psi}(0)=\mc{C}$ and does:
\State For each active client $i\in \bar{\Psi}(k)$, set $\tilde{r}_i(k)=\hat{r}_{i,a_i}(\bar{t}_i)$ if $k=0$; else if $k\geq 1$,  decode $\tilde{r}_{i}(k)= \texttt{Dec}_{i,k}(\sigma_i(k))$, and compute
$$ \tilde{r}^{(U)}_i(k)=\tilde{r}_i(k)+ 2\alpha_i(\bar{t}_i+kH); \hspace{2mm} \tilde{r}^{(L)}_i(k)=\tilde{r}_i(k) - 2\alpha_i(\bar{t}_i+kH)$$
\State \label{line:algo2line2} Compute  $\tilde{r}^{(L)}_{\max}(k) = \max_{i\in\bar{\Psi}(k)} \tilde{r}^{(L)}_{i}(k)$
\State Update active client set: $\bar{\Psi}(k+1) = \{i\in\bar{\Psi}(k) \ : \ \tilde{r}^{(L)}_{\max}(k) < \tilde{r}^{(U)}_{i}(k) \}$
\State \textbf{If} $|\bar{\Psi}(k+1)|>1$, broadcast the threshold $\tilde{r}^{(L)}_{\max}(k)$. \textbf{Else}, output $\bar{a} = a_{\bar{\Psi}(k+1)}$ and {terminate}
\Statex \hspace{-3mm} $\bullet$ Each active client $i \in \bar{\Psi}(k)$ does:
\State \textbf{If} $\tilde{r}^{(L)}_{\max}(k)$ is received from server \textit{and} $\tilde{r}^{(L)}_{\max}(k)  < \tilde{r}^{(U)}_{i}(k)$, then sample arm $a_i$ $H$ times and transmit  $\sigma_i(k+1)=\texttt{Enc}_{i,k+1}\left(\hat{r}_{i,a_i}(\bar{t}_i+(k+1)H)\right)$ to server. \textbf{Else}, deactivate 
\EndFor
\end{algorithmic}
 \end{algorithm}

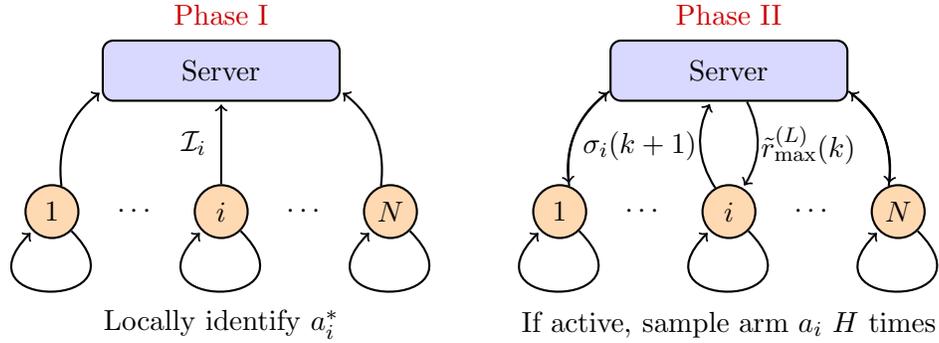
\begin{figure}[t]
\begin{center}
\begin{tikzpicture}
[->,shorten >=1pt,scale=.75,inner sep=1pt, minimum size=20pt, auto=center, node distance=3cm,
  thick, node/.style={circle, draw=black, thick},]
\tikzset{every loop/.style={min distance=10mm,in=225,out=315,looseness=7.5}}
\tikzstyle{block1} = [rectangle, draw, fill=red!10, 
    text width=8em, text centered, rounded corners, minimum height=0.8cm, minimum width=1cm];
\path[->, thick, blue];
\node [block1,draw, fill=blue!15](n1) at (0,2.5)
(s) {Server};
\node [circle, draw, fill=orange!30](n2) at (-3,0)     (1)  {$1$};
\node [circle, draw, fill=orange!30](n3) at (0,0)     (i)  {$i$};
\node [circle, draw, fill=orange!30](n4) at (3,0)     (N)  {$N$};
\node [black] at (-1.5,0) () {$\cdots$};
\node [black] at (1.5,0) () {$\cdots$};
\node at (0,-2) {Locally identify $a^*_i$};
\draw [->, thick] (1) to [bend left=30] (s);
\draw [->, thick] (i) to node [left] {$\mc{I}_i$} 
 (s);
\draw [->, thick] (N) to [bend right=30] (s);
\draw [->, thick] (i) to [loop below] (i);
\draw [->, thick] (1) to [loop below] (1);
\draw [->, thick] (N) to [loop below] (N);

\node [block1,draw, fill=blue!15](n1) at (9,2.5)
(s1) {Server};
\node [circle, draw, fill=orange!30](n2) at (6,0)     (2)  {$1$};
\node [circle, draw, fill=orange!30](n3) at (9,0)     (i1)  {$i$};
\node [circle, draw, fill=orange!30](n4) at (12,0)     (N1)  {$N$};
\node [black] at (7.5,0) () {$\cdots$};
\node [black] at (10.5,0) () {$\cdots$};
\node at (9,-2) {If active, sample arm $a_i$ $H$ times};
\node at (0,3.5) {\textcolor{winered}{Phase I}};
\node at (9,3.5) {\textcolor{winered}{Phase II}};
\draw [->, thick] (2) to [bend left=30] (s1);
\draw [->, thick] (i1) to [bend left=30] node [left] {$\sigma_i(k+1)$} 
 (s1);
\draw [->, thick] (N1) to [bend right=30] (s1);
\draw [->, thick] (i1) to [loop below] (i1);
\draw [->, thick] (2) to [loop below] (2);
\draw [->, thick] (N1) to [loop below] (N1);
\draw [->, thick] (s1) [bend left=-30] to (2);
\draw [->, thick] (s1) to [bend left=30] node [right] {$\tilde{r}^{(L)}_{\max}(k)$} (i1);
\draw [->, thick] (s1) to [bend right=-30] (N1);

\end{tikzpicture}
\end{center}
\caption{Illustration of \texttt{Fed-SEL}. (\textbf{Left}) In Phase I, each client $i$ performs local computations (represented by self-loops) to identify the best arm $a^*_i$ in $\mc{A}_i$. It then transmits $\mc{I}_i=\{a_i,\hat{r}_{i,a_i}(\bar{t}_i),\bar{t}_i\}$ to the server. (\textbf{Right}) In round $k$ of Phase II, the server broadcasts a threshold $\tilde{r}^{(L)}_{\max}(k)$ that is used by each client $i$ to determine if it is active. If active, it performs $H$ local sampling steps, and transmits an encoded version $\sigma_i(k+1)$ of the estimate $\hat{r}_{i,a_i}(\bar{t}_i+(k+1)H)$ of arm $a_i$'s mean.}
\label{fig:FedSEL}
\end{figure}

\textbf{Description of \texttt{Fed-SEL}:} Our proposed algorithm \texttt{Fed-SEL}, outlined in Algo.  \ref{algo:algo2}, and depicted in Fig. \ref{fig:FedSEL}, involves two distinct phases. In Phase I, each client $i\in\mathcal{C}$ runs a local successive elimination sub-routine (Algo. \ref{algo:algo1}) in parallel on its arm-set $\mathcal{A}_i$ to identify the locally best arm $a^*_i$ in $\mathcal{A}_i$. Note that this requires no communication. Phase II then involves periodic encoded communication between the clients and the server to identify the best arm among the locally best arms. We assume that each client $i$  knows the parameters $|\mc{C}|, |\mc{A}_i|$, and $\delta$, and that the server knows $|\mc{C}|, \{|\mc{A}_i|\}_{i\in\mc{C}}$, and $\delta$. Additionally, we assume that the communication period $H$ is known to both the clients and the server. We now describe each of the main components of \texttt{Fed-SEL} in detail.

\textbf{Phase I:} Algo. \ref{algo:algo1} is based on the successive elimination technique \cite{even}, and proceeds in epochs. In each epoch $t$, client $i$ maintains a set of active arms $\mathcal{S}_i(t) \subseteq \mathcal{A}_i$.  Each active arm $\ell \in \mathcal{S}_i(t)$ is sampled once, and its empirical mean $\hat{r}_{i,\ell}(t)$ is then updated. For the next epoch, an arm is retained in the active set if and only if its empirical mean is not much lower than the empirical mean of the arm with the highest empirical mean in epoch $t$; see line 5 of Algo. \ref{algo:algo1}. This process continues till only one arm is left, at which point Algo. \ref{algo:algo1} terminates at client $i$ and outputs the last arm $a_i$. Let $\bar{t}_i$ represent the final epoch count when Algo. \ref{algo:algo1} terminates at client $i$. It is easy to then see that arm $a_i$, the output of Algo. \ref{algo:algo1} (if it terminates) at client $i$, is sampled  $\bar{t}_i$ times at the end of Phase I. The quantities  $\{\alpha_i(t),\beta_i(t)\}$ associated with arm-set $\mathcal{A}_i$ are given by line 1 of Algo. \ref{algo:algo1}. Here, $c$ is a flexible parameter we introduce to control the accuracy of client $i$'s estimate $\hat{r}_{i,a_i}(\bar{t}_i)$.

\textbf{Phase II:} The goal of Phase II is to identify the best arm from the set {$\{a_i\}_{i\in\mathcal{C}}$}. Comparing arms across clients entails communication via the server. Since such communication is costly, we consider \textit{periodic exchanges} between the clients and the server as in standard FL algorithms. To further reduce communication, such exchanges are \textit{encoded} using a novel \textit{adaptive encoding} strategy that we will explain shortly. Specifically, Phase II operates in rounds, where in each round $k\in\{0,1,\ldots\}$, the server maintains a set of \textit{active clients} $\bar{\Psi}(k)$: a client $i$ is deemed active if and only if arm $a_i$ is still under contention for being the globally best arm. To decide upon the latter, the server aggregates the encoded information received from clients to compute the threshold $\tilde{r}^{(L)}_{\max}(k)$ (lines 3 and 4 of Algo. \ref{algo:algo2}). Computing this threshold requires some care to account for (i)  imbalances in sampling across arm-sets, and (ii) errors introduced due to encoding (quantization). Having computed $\tilde{r}^{(L)}_{\max}(k)$, the server executes line 5 of Algo. \ref{algo:algo2} to eliminate sub-optimal arms from the set $\{a_i\}_{i\in\bar{\Psi}(k)}$. It then updates the active client set and broadcasts $\tilde{r}^{(L)}_{\max}(k)$ if $|\bar{\Psi}(k+1)|>1$.

Only the arms that remain in $\{a_i\}_{i\in\bar{\Psi}(k+1)}$ require further sampling by the corresponding clients. Each client $i \in \bar{\Psi}(k)$ uses the threshold $\tilde{r}^{(L)}_{\max}(k)$ to determine whether it should sample $a_i$ any further (i.e., it checks whether to remain active or not). If the conditions in line 7 of Algo. \ref{algo:algo2} pass, then client $i$ samples arm $a_i$ $H$ times, and transmits an encoded version $\sigma_i(k+1)$ of $\hat{r}_{i,a_i}(\bar{t}_i+(k+1)H)$ to the server. Here, $H$ is the communication period; if $H=1$, then the active clients interact with the server at each time-step in Phase II. Since we aim to reduce the number of communication rounds, we instead let each client perform multiple local sampling steps before communicating with the server, i.e., $H > 1$ for our setting.\footnote{The local sampling steps in line 7 of \texttt{Fed-SEL} can be viewed as the analog of performing multiple local gradient-descent steps in standard federated optimization algorithms.} Phase II lasts  till the active client set reduces to a singleton, at which point the locally best arm corresponding to the last active client is deemed to be the globally best arm by the server. We now describe our encoding-decoding strategy.

\textbf{Adaptive Encoding Strategy:} At the end of round $k-1$, each active client $i\in\bar{\Psi}(k)$ encodes $\hat{r}_{i,a_i}(\bar{t}_i+kH)$ using a quantizer with range $[0,1]$ and bit-precision level $B_i(k)$, where 
\begin{equation}
    B_i(k)=\ceil*{\log_2 \left(\frac{1}{\alpha_i(\bar{t}_i+kH)}\right)}.
\label{eqn:bit_prec}
\end{equation}
Specifically, the interval $[0,1]$ is quantized into $2^{B_i(k)}$ bins, and each bin is assigned a unique binary symbol. Since all rewards belong to $[0,1]$, $\hat{r}_{i,a_i}(\bar{t}_i+kH)$ falls in one of the bins in $[0,1]$. Let $\sigma_i(k)$ be the binary representation of the index of this bin. Compactly, we use $\texttt{Enc}_{i,k}\left(\hat{r}_{i,a_i}(\bar{t}_i+kH)\right) =\sigma_i(k)$ to represent the above encoding operation at client $i$. The key idea we employ in our encoding technique described above is that of \textit{adaptive quantization}: the bit precision level $B_i(k)$ is \textit{successively refined} in each round  based on the confidence interval $\alpha_i(\bar{t}_i+kH)$.

For correct decoding, we assume that the server is aware of the encoding strategies at the clients, and the parameters  $|\mc{C}|, \{|\mathcal{A}_i|\}_{i\in\mathcal{C}}$, $H$,  and $\delta$. Based on this information, and the fact that at the end of Phase I, each client $i$ transmits $\bar{t}_i$ to the server, note that the server can \textit{exactly} compute $\alpha_i(\bar{t}_i+kH)$, and hence $B_i(k)$ for each $i\in\mc{C}$. Let $\tilde{r}_{i}(k)$ be the center of the bin associated with $\sigma_i(k)$. Then, we compactly represent the decoding operation at the server (corresponding to client $i$) as $\texttt{Dec}_{i,k}(\sigma_i(k))= \tilde{r}_{i}(k)$. Using $\tilde{r}_{i}(k)$ as a proxy for $\hat{r}_{i,a_i}(\bar{t}_i+kH)$, the server computes upper and lower estimates, namely $\tilde{r}^{(U)}_i(k)$ and $\tilde{r}^{(L)}_i(k)$, of the mean of arm $a_i$ in line 3 of Algo. \ref{algo:algo2}.

This completes the description of \texttt{Fed-SEL}. A few remarks are now in order. 

\begin{remark}
\texttt{Fed-SEL} is communication-efficient by design: Phase I requires incurs only one round of communication (at the end of the phase) while Phase II involves periodic communication. Moreover, in each round $k\in\{1,2,\ldots\}$ of Phase II, every client transmits encoded information about the empirical mean of only one arm from its arm-set. We note here that the encoding strategy needs to be designed carefully so as to prevent the  globally best-arm from being eliminated due to quantization errors; our proposed approach takes this into consideration. 
\end{remark}

\begin{remark}
In our current approach, the clients transmit information without encoding only once, at the end of Phase I (i.e., when $k=0$). This is done to inform the server of the sample counts $\{\bar{t}_i\}_{i\in\mc{C}}$ needed for correct decoding during Phase II.\footnote{Recall that in the encoding operation at the end of round $k-1$, the bit-precision $B_i(k)$ is a function of $\alpha_i(\bar{t}_i+kH)$. Other than $\bar{t}_i$, all other terms featuring in $\alpha_i(\bar{t}_i+kH)$ are known a priori to the server.} Below, we outline a simple way to bypass the need for transmitting the parameters $\{\bar{t}_i\}_{i\in\mc{C}}$. 

At the end of Phase I, suppose the first time the server receives information from client $i$ is at time-step $t'_i$.\footnote{Here, we implicitly assume that there are no transmission delays, and that the server is equipped with the means to keep track of time.}   Since client $i$ can sample only one arm at each time-step, the maximum number of epochs at client $i$ during Phase I is $t'_i/|\mc{A}_i| \triangleq \tilde{t}_i$. Thus, $\tilde{t}_i$ provides a lower bound on $\bar{t}_i$ - the number of times arm $a_i$ is sampled during Phase I. Using $\tilde{t}_i$ now in place of $\bar{t}_i$, one can employ  more conservative confidence intervals $\alpha_i(\tilde{t}_i+kH)$ both in line 3 of \texttt{Fed-SEL}, and also for encoding. While this strategy will also lead to identification of the best arm, it will in general require more rounds to terminate relative to \texttt{Fed-SEL}. As a final comment, using $O\left(\log_2{|\mc{A}_i|}\right)$ bits, one can encode the arm label $a_i$ as well. Based on the above discussion, it is easy to see that with minor modifications to our current algorithm, one can effectively encode all transmissions from the clients to the server. 
\end{remark}

\section{Performance Guarantees for \texttt{Fed-SEL}}
\label{sec:results_FedSEL}
\subsection{A Notion of Heterogeneity}
In Section \ref{sec:problem}, we provided an intuitive argument that statistical heterogeneity of arm distributions across different arm-sets should facilitate the task of identifying the globally best arm. To make this argument precise, however, we need to fill in two important missing pieces: (i) defining an appropriate notion of heterogeneity for our problem, and (ii) formally establishing that our proposed algorithm \texttt{Fed-SEL} benefits from the notion of heterogeneity so defined. In this section, we address both these objectives.  We start by introducing the concept of \textit{arm-heterogeneity indices} that will play a key role in our subsequent analysis of \texttt{Fed-SEL}. 

\begin{figure}[h]
\centering
  \includegraphics[width=0.5\linewidth]{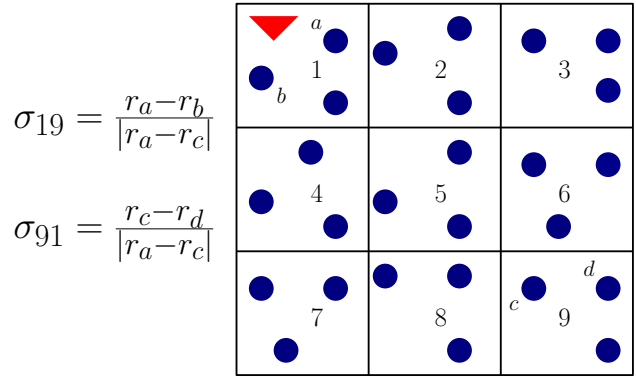}
  \caption{Illustration of Definition \ref{defn:arm_het} via a target detection example. An environment is partitioned into 9 sub-regions (arm-sets) with 3 sensors (arms) in each sub-region. The goal is to detect a target based on the sensor readings. For this purpose, an agent (client) is assigned to each sub-region. The target is depicted by the red triangle and the sensors are depicted by the blue circles. Arms $a$ and $b$ (resp., arms $c$ and $d$) are the two top arms (i.e., closest to the target) in sub-region 1 (resp., in sub-region 9). The arm-heterogeneity indices $\{\sigma_{19},\sigma_{91}\}$ for $\mc{A}_1$ and $\mc{A}_9$ are as shown in the figure. 
  }
  \label{fig:het}
\end{figure} 

\begin{definition} 
\label{defn:arm_het}
(\textbf{Arm-Heterogeneity Index}) For any two arm-sets $\mc{A}_i$ and $\mc{A}_j$ such that $r_{a^{(i)}_1} \neq r_{a^{(j)}_1}$, the arm-heterogeneity index $\sigma_{ij}$ is defined as follows:
\begin{equation}
    \sigma_{ij} = \frac{r_{a^{(i)}_1}-r_{a^{(i)}_2}}{\vert r_{a^{(i)}_1}-r_{a^{(j)}_1}\vert}.
\label{eqn:arm_het}
\end{equation}
\end{definition}

With Definition \ref{defn:arm_het}, we aim to capture the intuitive idea that \textit{distributions of arms within an arm-set are more similar than those across arm-sets}. The arm-heterogeneity indices we have introduced quantify the extent to which the above statement is true. In words, $\sigma_{ij}$ measures the closeness between the two best arms within $\mc{A}_i$ relative to the closeness between the best arms in each of the sets $\mc{A}_i$ and $\mc{A}_j$; here, ``closeness" between two arms is quantified by the difference in their means.\footnote{Recall that we use $a^{(i)}_1$ to denote the top arm in $\mc{A}_i$, i.e., the arm with the highest mean reward in $\mc{A}_i$.} Together, the index pair $\{\sigma_{ij},\sigma_{ji}\}$ captures the level of dissimilarity or heterogeneity between arm-sets $\mc{A}_i$ and $\mc{A}_j$; smaller values of these indices reflect more heterogeneity. To see why the latter is true, notice that both $\sigma_{ij}$ and $\sigma_{ji}$ scale inversely with the gap in mean rewards of the two best arms in $\mc{A}_i$ and $\mc{A}_j$. Thus, if these arms have ``well-separated" mean rewards - an intuitive measure of heterogeneity - then this would lead to small values of $\sigma_{ij}$ and $\sigma_{ji}$. 

\textbf{How does heterogeneity arise in practice?}  Our definition of heterogeneity is motivated by scenarios where the partitioning of arms into disjoint arm-sets is not arbitrary, but rather adheres to some structure. Consider for instance the target detection application we discussed in the Introduction; see Fig. \ref{fig:het} for an illustration. For this setting, sensors closer to the target have higher readings than those farther away. As a result, one should expect the readings of sensor $a$ to be more similar to those of sensor $b$ (in the same sub-region) as compared to those of sensor $c$ in sub-region 9. This \textit{spatial heterogeneity} effect is precisely captured by low values of the arm-heterogeneity index $\sigma_{19}.$ One can similarly argue that spatial heterogeneity leads to low values of $\sigma_{91}$.

\subsection{Main Result and Discussion}
Our first main result, Theorem \ref{thm:Fed-SEL}, shows that \texttt{Fed-SEL} is $(0,\delta)$-PAC, and provides a detailed characterization of its sample- and communication-complexity. To state the result cleanly, we will use $i_1$ as a shorthand for $a^{(i)}_1$, and employ  the following notations:
$$
\bar{c}_i = \sqrt{\frac{4|\mc{C}| |\mc{A}_i|}{\delta}}, \hspace{2mm}  \Delta^*_i=r_{a^{(i)}_1}-r_{a^{(i)}_2}, \hspace{2mm} \textrm{and} \hspace{2mm} \Delta^*= r_1 - \max_{i\in\mc{C}\setminus\{1\}} r_{i_1}. 
$$
Thus, $\Delta^*$ is the difference in the mean rewards of the two best arms in $\mc{A}$. We remind the reader here that the best arm $a^*=1$ belongs to the arm-set $\mc{A}_1$ of client $1$. 

\begin{theorem}
\label{thm:Fed-SEL}
Suppose $\texttt{Fed-SEL}$ is run with $\beta_i(t)=8\alpha_i(t), \forall i\in\mc{C}$. Then with probability at least $1-\delta$, the following statements hold.

\begin{itemize}
\item \textbf{(Consistency)}  \texttt{Fed-SEL} terminates in finite-time and outputs the globally best arm, i.e., we have $\bar{a}=a^*$.

\item \textbf{(Communication-Complexity)} The number of rounds $R_i$ a client $i\in\mc{C}\setminus\{1\}$ remains active during Phase II is bounded above as follows: $R_i \leq \max\{R_{i1},R_{1i}\}$, where 
\begin{equation}
    R_{i1} \leq \frac{1}{H}\left( 
    \frac{128}{{(\Delta^*_i)}^2} (\sigma^2_{i1} -1) \log{\frac{128\bar{c}_i}{\Delta^2_{1i_1}}} + \frac{256}{{(\Delta^*_i)}^2} \log{\sigma_{i1}} + O\left(\frac{1}{\Delta^2_{1i_1}}\log\log{\frac{\bar{c}_i}{\Delta^2_{1i_1}}} \right) \right)+1, \hspace{1mm} \textrm{if} \hspace{1mm} \sigma_{i1} > 1,
\label{eqn:Ri1}
\end{equation}
and $R_{i1}=1$ if $\sigma_{i1} \leq 1$. Similarly, 
\begin{equation}
    R_{1i} \leq \frac{1}{H}\left( 
    \frac{128}{{(\Delta^*_1)}^2} (\sigma^2_{1i} -1) \log{\frac{128\bar{c}_1}{\Delta^2_{1i_1}}} + \frac{256}{{(\Delta^*_1)}^2} \log{\sigma_{1i}} + O\left(\frac{1}{\Delta^2_{1i_1}}\log\log{\frac{\bar{c}_1}{\Delta^2_{1i_1}}} \right) \right)+1, \hspace{1mm} \textrm{if} \hspace{1mm} \sigma_{1i} > 1,
\label{eqn:R1i}
\end{equation}
and $R_{1i}=1$ if $\sigma_{1i} \leq 1$. The number of communication rounds $R$ during Phase II is then given by $R=\max_{i\in\mc{C}\setminus\{1\}}{R_i}$;  client $1$ always remains active till \texttt{Fed-SEL} terminates, i.e., $R_1=R$. Moreover, each client $i\in\mc{C}$ transmits  at most
$$  O\left(\log_2{\left(\frac{1}{\Delta^*}\right)}\right) $$
bits of information in each round $k\in\{1,\ldots, R_{i1}-1\}$. 

\item (\textbf{Sample-Complexity}) The total number of arm-pulls made by client $i$ is bounded above by
\begin{equation}
\underbrace{O\left( \sum_{\ell \in \mathcal{A}_i \setminus \{i_1\}} \frac{\log\left(\frac{|\mc{C}| |\mc{A}_i|}{\delta \Delta_{i_1 \ell}}\right)}{\Delta^2_{i_1 \ell}} \right)}_{\textrm{Phase-I arm-pulls}} + \underbrace{HR_i}_{\textrm{Phase-II arm-pulls}}\hspace{-8mm}.
\label{eqn:arm_pulls}
\end{equation}
\end{itemize}
\end{theorem}

Since each client can only sample arms from its own arm-set, it is apparent that at least one round of communication is necessary for solving Problem \ref{prob:FedMAB}, in general. The next result - an immediate corollary of Theorem \ref{thm:Fed-SEL} -  identifies heterogeneous regimes where one round of communication is also sufficient.\footnote{When $\sigma_{i1}=1$, we would ideally like the upper-bound on ${R}_{i1}$ to evaluate to $1$.  Setting $\sigma_{i1}=1$ in \eqref{eqn:Ri1} causes the first two dominant terms in the upper-bound for ${R}_{i1}$ to vanish, as desired. The lower order term that remains is an artifact of our analysis.}  

\begin{corollary}
\label{corr:oneround}
Suppose $\texttt{Fed-SEL}$ is run with $\beta_i(t)=8\alpha_i(t), \forall i\in\mc{C}$. Additionally, suppose $$\max_{i\in\mc{C}\setminus\{1\}} \{\sigma_{1i}, \sigma_{i1}\} \leq 1.$$ Then, with probability at least $1-\delta$, \texttt{Fed-SEL} terminates after just one round of communication between the clients and the server, and outputs the globally best arm, i.e., $\bar{a}=a^*$. 
\end{corollary}

The proof of Theorem \ref{thm:Fed-SEL} is deferred to Appendix \ref{app:thm1proof}. In what follows, we discuss several important implications of {Theorem \ref{thm:Fed-SEL}}. 

$\bullet$ (\textbf{Benefit of Heterogeneity}) From the first two dominant terms in the expressions for $R_{i1}$ and $R_{1i}$ in equations \eqref{eqn:Ri1} and \eqref{eqn:R1i}, respectively, observe that lower  values of the arm-heterogeneity indices translate to fewer communication rounds. Moreover, fewer rounds imply fewer arm-pulls (and hence, lower sample-complexity)  during Phase II, as is evident from  Eq. \eqref{eqn:arm_pulls}. Since low  values of arm-heterogeneity indices correspond to high  heterogeneity, we conclude that \textit{heterogeneity helps reduce both the sample- and communication-complexity of \texttt{Fed-SEL}}. 

$\bullet$ (\textbf{Benefit of Local Steps}) Even if the level of heterogeneity is low in a given instance, one can reduce the number of communication rounds of \texttt{Fed-SEL} by increasing the communication period $H$; see equations \eqref{eqn:Ri1} and \eqref{eqn:R1i}. Based on these equations and \eqref{eqn:arm_pulls}, we note that $H$ shows up as an additive term in the upper-bound on the number of arm-pulls made during Phase II. The key implication of this fact is that \textit{one can ramp up $H$ to significantly reduce communication and get away with a modest price in terms of additional sample-complexity.} The main underlying  reason for this stems from the observation that with  \texttt{Fed-SEL}, more local steps always help by providing better arm-mean estimates. 

$\bullet$ (\textbf{Benefit of Collaboration}) We have already discussed how heterogeneity assists in reducing the sample-complexity of Phase II. To focus on the number of arm-pulls made by a client $i$ during Phase I, let us now turn our attention to the first term in Eq. \eqref{eqn:arm_pulls}. The summation in this term is over the local arm-set $\mc{A}_i$ of client $i$, as opposed to the global arm-set $\mc{A}$. Since $|\mc{A}_i|$ can be much smaller than $|\mc{A}|$ in practice, the benefit of collaboration is immediate: each client needs to explore only a small subset of the global arm-set, i.e., \textit{the task of exploration is distributed among clients}. 

$\bullet$ (\textbf{Number of Bits Exchanged}) Since $\Delta^*$ is the gap in mean rewards of the two best arms (i.e., the two arms that are hardest to distinguish) in the global arm set $\mc{A}$, the quantity $\log_2{\left(1/\Delta^*\right)}$ essentially captures the complexity of a given instance of the MAB problem. Intuitively, one should expect that the amount of information that needs to be exchanged to solve a given instance of Problem \ref{prob:FedMAB} should scale with the complexity of that instance, i.e., with $\log_2{\left(1/\Delta^*\right)}$. From Theorem \ref{thm:Fed-SEL}, we note that our adaptive encoding strategy formalizes this intuition to a certain extent.

\begin{remark} From the above discussion, note that although our algorithm itself is oblivious to both the level of heterogeneity and the complexity of a given instance, the sample- and communication-complexity bounds we obtain naturally adapt to these defining features of the instance, i.e., our guarantees are instance-dependent.
\end{remark}

We now briefly comment on the main steps in the proof of Theorem \ref{thm:Fed-SEL}. 

\textbf{Proof Sketch for Theorem \ref{thm:Fed-SEL}}:  We first define a ``good event" $\mc{E}$ with measure at least $1-\delta$, on which,  the empirical means of all arms are within their respective confidence intervals at all times. On this event $\mc{E}$, we show using standard arguments that at the end of Phase I, each client $i$ successfully identifies the locally best arm $a^*_i$ in $\mc{A}_i$, i.e., $a_i=a^*_i=a^{(i)}_1$. At this stage, to exploit heterogeneity, we require the locally best arm from each arm-set to be ``well-explored". Accordingly, we  establish that $\vert\hat{r}_{i,a_i}(\bar{t}_i)-r_{a^{(i)}_1}\vert \leq c/{(c-2)}^2 \Delta^*_i, \forall i\in\mc{C}$, where $c$ is the flexible design parameter in line 1 of Algo. \ref{algo:algo1}. The next key piece of our analysis pertains to relating the amount of exploration that has already been done during Phase I, to that which remains to be done during Phase II for eliminating all arms in the set $\{a^{(i)}_1\}_{i\in\mc{C}\setminus\{1\}}$; this is precisely where the arm-heterogeneity indices show up in the analysis. To complete the above step, we derive new upper and lower bounds on the Lambert $W$ function that may be of independent interest; see Lemma \ref{lemma:lambert} in Appendix \ref{app:lambert_bnds}.  

Before moving on to the next section, we discuss how certain aspects of our algorithm and analysis can be potentially improved. 

\textbf{Potential Refinements:} As is most often the case, there is considerable room for improvement. First, the dependence on the number of arms within the logarithmic terms of our sample-complexity bounds can be improved using a variant of the Median Elimination Algorithm \cite{even2002}. Second, note from \eqref{eqn:arm_pulls} that for each client $i$, its sample-complexity during Phase I scales inversely with the  local arm-gaps $\Delta_{i_1\ell}$, where $\ell\in\mc{A}_i\setminus\{i_1\}$, and $i_1=a^{(i)}_1$ is the locally best-arm in $\mc{A}_i$. Ideally, however, we would like to achieve sample-complexity bounds that scale inversely with $\Delta_{1\ell}$, where $1\in\mc{A}_1$ is the globally best arm. For all clients other than client 1 (i.e., clients who cannot sample arm $1$ directly), this would necessarily require communication via the server. Thus, while we save on communication during Phase I, the price paid for doing so is a larger sample-complexity.

One may thus consider the alternative of merging Phases I and II. Our reasons, however, for persisting with separate phases are as follows. First, in our current approach, a client starts communicating when the best arm in its arm-set has already been well-explored. This, in turn, considerably reduces the workload for Phase II, and helps reduce communication significantly (see Corollary \ref{corr:oneround} for instance). The second important reason is somewhat more subtle: as we explain in the next section, having two separate phases aids the task of tolerating adversarial clients. 

\section{Robust Federated Best-Arm Identification}
\label{sec:robust_variant}
In this section, we will focus on the following question: \emph{Can we still hope to identify the best-arm when certain clients act adversarially?} To formally answer this question, we will consider a Byzantine adversary model \cite{pease,dolev} where adversarial clients have complete knowledge of the model, can collude among themselves, and act arbitrarily. Let us start by building some intuition. Suppose the client associated with the arm-set containing the globally best arm $a^*$ is adversarial. Since this is the only client that can sample $a^*$, there is not much hope of identifying the best arm in this case. This simple observation reveals the need for \textit{information-redundancy: we need multiple clients to sample each arm-set to account for adversarial behaviour.} Accordingly, with each arm-set $\mathcal{A}_j$, we now associate a group of clients $\mathcal{C}_j$; thus, $\mc{C}=\cup_{j\in [N]} \mc{C}_j$.\footnote{From now on, we will use the index $j$ to refer to arm-sets and client-groups, and the index $i$ to refer to clients.} We will denote the set of honest clients by $\mc{H}$, the set of adversarial clients by $\mc{J}$, and allow at most $f$ clients to be adversarial in each client-group, i.e., $|\mathcal{C}_j\cap \mathcal{J}| \leq f, \forall j\in\mathcal [N]$. We will also assume that the server knows $f$.

Simply restricting the number of adversaries in each client-group is not enough to identify the best arm; there are additional challenges that need to be dealt with. To see this, consider a scenario where an arm $\ell \in \mc{A}_j$ has not been adequately explored by an honest client $i\in\mc{C}_j$. Due to the stochastic nature of our model, client $i$, although honest, may have a poor estimate of arm $\ell's$ true mean reward at this stage. How can we distinguish this scenario from one where an adversarial client in $\mc{C}_j$ deliberately transmits a completely incorrect estimate of arm $\ell$? In this context, one natural idea is to enforce each honest client to start transmitting information only when it has sufficiently explored every arm in its arm-set. This further justifies the rationale behind having two separate phases in \texttt{Fed-SEL}. In what follows, we describe a robust variant of \texttt{Fed-SEL}. 

\begin{algorithm}[H]
\caption{Robust Federated Successive Elimination (\texttt{Robust Fed-SEL})}
\label{algo:algo3}  
 \begin{algorithmic}[1] 
\Statex \hspace{-5mm} \textbf{Phase I:} \textit{Identifying locally best arms}
\State For each $j\in [N]$, every client $i\in\mathcal{C}_j \cap \mc{H}$ runs Algo. \ref{algo:algo1} on $\mathcal{A}_j$ with $\beta_j(t)=8\alpha_j(t)$. If Algo. \ref{algo:algo1} terminates at client $i$, then client $i$ reports $\{a_i,\hat{r}_{i,a_i}(\bar{t}_i),\bar{t}_i\}$ to the server
\Statex \hrulefill
\Statex \hspace{-5mm} \textbf{Phase II:} \textit{Identifying the globally best arm}
\State For each $j \in [N]$, server identifies a  representative arm for $\mathcal{A}_j$ as $\bar{a}_j= \texttt{Maj\textunderscore Vot}\left(\{a_i\}_{i\in\mc{C}'_j}\right)$, where $\mc{C}'_j\subseteq \mc{C}_j$ are the first $(2f+1)$ clients from $\mc{C}_j$ to report to the server at the end of Phase I
\State Let  $\bar{\mc{C}}_j =\{i\in \mc{C}'_j: a_i = \bar{a}_j\}$ be those clients in $\mc{C}'_j$ that report $\bar{a}_j$, and define  $f_j = f-|\mc{C}'_j| + |\bar{\mc{C}}_j|$ 
 \For {$k\in\{0,1,\ldots\}$}
\Statex \hspace{-3mm} $\bullet$ Server initializes $\bar{\Psi}(0)=[N]$ and does:
\State For each active client-group $j\in \bar{\Psi}(k)$, decode  $\tilde{r}_i(k)$ and compute the pair $\{\tilde{r}^{(U)}_i(k),\tilde{r}^{(L)}_{i}(k)\}$ exactly as in \texttt{Fed-SEL} for every client $i\in\bar{\mc{C}}_j$ 
\State For each $j\in \bar{\Psi}(k)$, compute \textit{robust} upper and lower estimates of the mean reward of $\bar{a}_j$:
$$\bar{r}^{(U)}_j(k) = \texttt{Trim}\left( \{\tilde{r}^{(U)}_{i}(k)\}_{i\in\bar{\mc{C}}_j},f_j \right) \hspace{1mm} \textrm{and} \hspace{1mm} \bar{r}^{(L)}_j(k) = \texttt{Trim}\left( \{\tilde{r}^{(L)}_{i}(k)\}_{i\in\bar{\mc{C}}_j},f_j \right)$$
\State Compute $\bar{r}^{(L)}_{\max}(k)=\max_{j\in\bar{\Psi}(k)} \bar{r}^{(L)}_j(k)$
\State Update active client groups: $\bar{\Psi}(k+1) = \{j\in\bar{\Psi}(k) \ : \ \bar{r}^{(L)}_{\max}(k) < \bar{r}^{(U)}_{j}(k) \}$ 
\State \textbf{If} $|\bar{\Psi}(k+1)|>1$, broadcast a binary $N \times 1$ vector $\mb{d}_k$ that encodes active client-groups. \textbf{Else}, output $\bar{a} = \bar{a}_{\bar{\Psi}(k+1)}$ and {terminate}
\Statex \hspace{-3mm} $\bullet$ For each $j\in [N]$, each client $i \in \bar{\mathcal{C}}_j \cap \mc{H}$ does: 
\State \textbf{If}  $\mathbf{d}_k$ is received from server and $\mathbf{d}_k(j)=1$, then sample arm $a_i$ $H$ times, encode $\hat{r}_{i,a_i}(\bar{t}_i+(k+1)H)$ as in \texttt{Fed-SEL}, and transmit the encoding $\sigma_i(k+1)$ to the server. \textbf{Else}, deactivate
\EndFor
\end{algorithmic}
 \end{algorithm}

\textbf{Description of \texttt{Robust Fed-SEL:}} To identify $a^*$ despite adversaries, we propose \texttt{Robust Fed-SEL}, and outline its main steps in Algorithm \ref{algo:algo3}. Like the \texttt{Fed-SEL} algorithm, \texttt{Robust Fed-SEL} also involves two separate phases. During Phase I, in each client group $\mc{C}_j, j\in [N]$, every honest client $i\in \mc{C}_j \cap \mc{H}$ independently runs Algo. \ref{algo:algo1} on $\mathcal{A}_j$, and then transmits $\{a_i,\hat{r}_{i,a_i}(\bar{t}_i),\bar{t}_i\}$ to the server  exactly as in \texttt{Fed-SEL}. The key operations performed by the server as follows. 

\textbf{1) Majority Voting to Identify Representative Arms:} At the onset of Phase II, the server waits till it has heard from at least $(2f+1)$ clients of each client group before it starts acting. Next, for each arm-set $\mathcal{A}_j$, the server aims to compute a \textit{representative locally best arm}  $\bar{a}_j$. Note that adversarial clients in $\mathcal{C}_j$ may collude and decide to report an arm to the server that has true mean much lower than that of the locally best arm $a^{(j)}_1$ in $\mathcal{A}_j$. Nonetheless, we should expect every honest client $i\in\mathcal{C}_j$ to output $a_i=a^{(j)}_1$ with high probability. Thus, if honest clients are in majority in $\mathcal{C}_j$, a simple majority voting operation (denoted by \texttt{Maj\_Vot}) at the server should suffice to reveal the best arm in $\mathcal{A}_j$ with high probability; this is precisely the rationale behind line 2. During the remainder of Phase II, for each client group $j\in[N]$, the server only uses the information of the subset $\bar{\mc{C}}_j \subset \mc{C}_j$ of clients who reported the representative arm $\bar{a}_j$. 

\textbf{2) Trimming to Compute Robust Arm-Statistics:} Phase II involves periodic communication and evolves in rounds. In each round $k$, the server maintains a set $\bar{\Psi}(k)$ of active client-groups: group $j$ is deemed active if $\bar{a}_j$ is still under contention for being the globally best arm. To decide whether group $j$ should remain active, the server performs certain trimming operations that we next describe. Given a non-negative integer $e$, and a set $\{m_i\}_{i\in[K]}$ of $K$ scalars where $K \geq 2e+1$, let $\texttt{Trim}\left(\{m_i\}_{i\in[K]}, e\right)$ denote the operation of removing the highest $e$ and lowest $e$ numbers from  $\{m_i\}_{i\in[K]}$, and then returning any one of the remaining numbers. Based on this \texttt{Trim} operator, the server computes \textit{robust upper and lower estimates, namely $\bar{r}^{(U)}_j(k)$ and $\bar{r}^{(L)}_j(k)$, of the mean of the representative arm $\bar{a}_j, \forall j\in\bar{\Psi}(k) $}; see line 6.\footnote{The trimming parameter $f_j$ is set based on the observation that clients in $\mc{C}'_j$ who report an arm other than $\bar{a}_j$ are adversarial with high probability.} Using these robust estimates, the server subsequently updates the set of active client-groups by executing lines 7-8.

If group $j$ remains active, then the server needs more refined arm estimates of arm $\bar{a}_j$ from $\bar{\mc{C}}_j$. Accordingly, the server sends out a binary $N \times 1$ vector $\mb{d}_k$ that encodes the active client-groups: $\mb{d}_k(j)=1$ if and only if $j\in\bar{\Psi}(k+1)$. Upon receiving $\mb{d}_k$, each honest client $i$ in $\bar{\mc{C}}_j$ checks if $\mb{d}_k(j)=1$. If so, it samples arm $a_i = \bar{a}_j$ $H$ times, encodes $\hat{r}_{i,a_i}(\bar{t}_i+(k+1)H)$ as in \texttt{Fed-SEL}, and transmits the encoding $\sigma_i(k+1)$ to the server. If $\mb{d}_k$ is not received from the server, or if $\mb{d}_k(j)=0$, client $i$ does nothing further. This completes the description of \texttt{Robust Fed-SEL}.

\textbf{Comments on Algorithm \ref{algo:algo3}:} Before providing formal guarantees for \texttt{Robust Fed-SEL}, a few comments are in order. First, note that unlike \texttt{Fed-SEL} where the server transmitted a threshold to enable the clients to figure out whether they are active or not, the server instead transmits a $N \times 1$ vector $\mb{d}_k$ in \texttt{Robust Fed-SEL}. The reason for this is as follows. Even if $\bar{r}^{(L)}_{\max}(k) \geq \bar{r}^{(U)}_{j}(k)$ for some client-group $j$, it may very well be that $\bar{r}^{(L)}_{\max}(k) < \tilde{r}^{(U)}_{i}(k)$ for some $i\in\bar{\mc{C}}_j \cap \mc{H}.$ Thus, even if the server is able to eliminate client group $j$ on its end, simply broadcasting the threshold $\bar{r}^{(L)}_{\max}(k)$ may not be enough for every client in $\bar{\mc{C}}_j \cap \mc{H}$ to realize that it no longer needs to sample arm $\bar{a}_j$. Note that for line 10 of \texttt{Robust Fed-SEL} to make sense, we assume that every client is aware of the client-group to which it belongs. 

During Phase II, for each group $j$, since the server only uses the information of clients in $\bar{\mc{C}}_j$ , we have not explicitly specified actions for honest clients in $\mc{C}_j \setminus \bar{\mc{C}}_j$ to avoid cluttering the exposition. A potential specification is as follows. Suppose the first time an honest client $i \in \mc{C}_j$ hears from the server, it is still in the process of completing Phase I. Since in this case client $i$ cannot belong to $\bar{\mc{C}}_j$, it deactivates and does nothing further. In all other cases, client $i \in \mc{C}_j$ acts exactly as specified in line 10 of \texttt{Robust Fed-SEL}. 

The main result of this section is as follows.

\begin{theorem}
\label{thm:rFedSEL}
Suppose $|\mathcal{C}_j\cap \mathcal{J}| \leq f$ and $|\mathcal{C}_j| \geq 3f+1, \forall j\in\mathcal [N]$. Then, \texttt{Robust Fed-SEL} is $(0,\delta)$-PAC.
\end{theorem}

 \textbf{Discussion:} Under information-redundancy, i.e., under the condition that  ``enough" clients sample every arm in each arm-set ($|\mathcal{C}_j| \geq 3f+1, \forall j\in\mathcal [N]$), Theorem \ref{thm:rFedSEL} shows that 
 \texttt{Robust Fed-SEL} is an adversarially-robust $(0,\delta)$-PAC algorithm.  Despite the flurry of research on tolerating adversaries in distributed optimization/supervised learning \cite{chensu,blanchard,dong,draco,alistarhbyz,xie,lirsa,ghosh1,ghosh2,ghosh3}, there is no analogous result (that we know of) in the context of MAB problems. Theorem \ref{thm:rFedSEL}  fills this gap. The sample- and communication-complexity bounds for \texttt{Robust Fed-SEL} are of the same order as that of \texttt{Fed-SEL}; hence, we do not explicitly state them here. For a discussion on this topic, see Appendix \ref{app:RFedSELproof} where we provide the proof of Theorem \ref{thm:rFedSEL}.

\begin{remark}
The type of redundancy assumption we make in Theorem \ref{thm:rFedSEL} is extremely standard in the literature on tolerating worst-case adversarial models. Similar assumptions are made in \cite{chensu,blanchard,dong,draco,alistarhbyz,xie,lirsa,ghosh1,ghosh2,ghosh3} where typically less than half of the agents (clients) are allowed to be adversarial. The reason why we allow at most $1/3$ of the clients in each client-group to be adversarial (as opposed to at most $1/2$) is because some adversaries may choose not to transmit anything at all. In this case, the server may never acquire enough information from $\mc{C}_j$ to compute $\bar{a}_j$. Now consider a slightly weaker adversary model where every client transmits when it is supposed to, but the information being sent can be arbitrarily corrupted for clients under attack. For this setting, our results go through identically under the weaker assumption that $|\mathcal{C}_j\cap \mathcal{J}| \leq f$ and $|\mathcal{C}_j| \geq 2f+1, \forall j\in\mathcal [N]$.
\end{remark} 
 
We now investigate how the ideas developed herein can be extended to the peer-to-peer setting.  
 \section{Robust Peer-to-Peer Best-Arm Identification}
\label{sec:MAB_network}
\begin{algorithm}[H]
\caption{Robust Networked Successive Elimination protocol at each Client $i\in\mathcal{C}_j\cap\mc{H}$}
\label{algo:p2p}  
 \begin{algorithmic}[1]
\Statex \hspace{-5 mm} \textbf{Step 1:} \textit{Updating $\mathbf{I}_i(j)$ and $\mathbf{R}_i(j)$}
\State Run Algo. \ref{algo:algo1} on $\mc{A}_j$ with $\beta_j(t)=6\alpha_j(t)$. If Algo. \ref{algo:algo1} terminates, set $\mathbf{I}_i(j)=a_i$, $\mathbf{R}_i(j)=\hat{r}_{i,a_i}(\bar{t}_i)$, and transmit $\mathbf{I}_i(j), \mathbf{R}_i(j)$ to each $k\in \mc{N}^{+}_i$
\Statex \hspace{-5mm} \textbf{Step 2:} \textit{Updating $\mathbf{I}_i(\ell)$ and  $\mathbf{R}_i(\ell), \forall \ell \in [N]\setminus\{j\}$}
\State Let $\mc{N}'_{i,\ell}\subseteq \mc{N}_i$ be the first $(2f+1)$ in-neighbors of $i$ that report about arm-set $\ell$
\State Set $\mathbf{I}_i(\ell)=\texttt{Maj\textunderscore Vot}\left(\{\mathbf{I}_k(\ell)\}_{k\in\mc{N}'_{i,\ell}}\right)$. Let $\bar{\mc{N}}_{i,\ell}=\{k\in\mc{N}'_{i,\ell}: \mathbf{I}_k(\ell)=\mathbf{I}_i(\ell)\}$ 
\State Set $\mathbf{R}_i(\ell)=\texttt{Trim}\left( \{\mathbf{R}_k(\ell)\}_{k\in\bar{\mc{N}}_{i,\ell}},f_{i,\ell} \right)$, where $f_{i,\ell} = f-|\mc{N}'_{i,\ell}| + |\bar{\mc{N}}_{i,\ell}|$. Transmit $\mathbf{I}_i(\ell), \mathbf{R}_i(\ell)$ to each $k\in \mc{N}^{+}_i$
\Statex \hspace{-5mm} \textbf{Step 3:} \textit{Identifying globally best arm}
\State Let $p^*_i=\argmax_{\ell\in [N]} \mathbf{R}_i[\ell]$. \textbf{Output} $\mathbf{I}_i(p^*_i)$. 
\end{algorithmic}
 \end{algorithm}
 
In this section, we consider the challenging setting of best-arm identification over a fully-distributed network in the presence of adversarial clients. Formally, the network is described by a static directed graph $\mc{G}=(\mc{V},\mc{D})$, where the vertex set $\mc{V}$ is composed of the set of clients, i.e., $\mc{V}=\mc{C}=\cup_{j\in[N]} \mc{C}_j$, and $\mc{D}$ represents the set of edges between them. An edge $(u,v)\in\mathcal{D}$ indicates that client $u$ can directly transmit information to client $v$. The set of all in-neighbors (resp., out-neighbors) of client $u$ is defined as $\mathcal{N}_u=\{v\in\mathcal{V}:(v,u)\in\mathcal{D}\}$ (resp., $\mathcal{N}^+_u=\{v\in\mathcal{V}:(u,v)\in\mathcal{D}\}$). 

Our \textbf{goal} is to design a distributed algorithm that ensures that each honest client $i\in \mc{H}$ can eventually identify the globally best arm $a^*$, despite the actions of the adversarial clients $\mc{J}$. Clearly, we need some upper-bound on the number of adversaries for the problem to make sense. To this end, we will consider an \textit{$f$-local Byzantine adversary model} \cite{flocal1,flocal2,rescons,mitraByz} where there are at most $f$ adversarial clients in the in-neighborhood of each honest client, i.e., $|\mc{N}_i \cap \mc{J}| \leq f, \forall i \in \mc{H}$.  Recall that clients in $\mathcal{C}_j$ can only sample arms from $\mc{A}_j$. Thus, to acquire information about some other arm-set $\mc{A}_{\ell}$ (say), we need information to diffuse from the clients in $\mathcal{C}_{\ell}$ to clients in $\mathcal{C}_{j}$. If there is a small cut  (bottleneck) in the graph separating $\mathcal{C}_{\ell}$ from $\mathcal{C}_{j}$, then such a cut can be infested by adversaries, thereby preventing the flow of information necessary for identifying $a^*$. From this intuitive reasoning, note that we need sufficient disjoint paths linking different client groups to have any hope of tackling adversaries. The key graph-theoretic property that we will exploit in this context is a notion of \textit{strong-robustness} \cite{mitraByz} adapted to our setting.  

\begin{figure}[t]
\begin{center}
\begin{tikzpicture}
[->,shorten >=1pt,scale=.75,inner sep=1pt, minimum size=20pt, auto=center, node distance=3cm,
  very thick, node/.style={circle, draw=black, thick},]
\tikzset{every loop/.style={min distance=10mm,in=225,out=315,looseness=7.5}}
\tikzstyle{block1} = [rectangle, draw, fill=red!10, 
    text width=8em, text centered, rounded corners, minimum height=0.8cm, minimum width=1cm];
\path[->, thick, blue];
\node [circle, draw,fill= gray!20](n5) at (0,0)     (5)  {$5$};

\node [circle, draw, fill=gray!20](n1) at (-6,3)        (1)  {$1$};
\node [circle, draw, fill=gray!20](n2) at (-4,1.5)      (2)  {$2$};
\node [circle, draw, fill=gray!20](n3) at (-4,-1.5)     (3)  {$3$};
\node [circle, draw, fill=gray!20](n4) at (-6,-3)       (4)  {$4$};

\node [circle, draw, fill=gray!20](n6) at (6,-3)        (6)  {$6$};
\node [circle, draw, fill=gray!20](n7) at (4,-1.5)      (7)  {$7$};
\node [circle, draw, fill=gray!20](n8) at (4,1.5)       (8)  {$8$};
\node [circle, draw, fill=gray!20](n9) at (6,3)         (9)  {$9$};

\node at (-5,4.75) {$\mc{C}_1$};
\node at (5,4.75) {$\mc{C}_3$};
\node at (0,1.5) {$\mc{C}_2$};

\draw [-, very thick] (1) to (2);
\draw [-, very thick] (1) to (3);
\draw [-, very thick] (1) to (4);
\draw [-, very thick] (2) to (3);
\draw [-, very thick] (2) to (4);
\draw [-, very thick] (3) to (4);

\draw [-, very thick] (6) to (7);
\draw [-, very thick] (6) to (8);
\draw [-, very thick] (6) to (9);
\draw [-, very thick] (7) to (8);
\draw [-, very thick] (7) to (9);
\draw [-, very thick] (8) to (9); 

\draw [-, very thick] (2) to [bend left=20] (5);
\draw [-, very thick] (3) to [bend right=20] (5);
\draw [-, very thick] (8) to [bend right=20] (5);
\draw [-, very thick] (7) to [bend left=20] (5);

\node (rect) at (-5,0) (c1) [draw, dashed, very thick, red, rounded corners,  minimum width=2.5cm, minimum height=6.25cm] {};
\node (rect) at (5,0) (c2) [draw, dashed, very thick, blue, rounded corners,  minimum width=2.5cm, minimum height=6.25cm] {};
\node (rect) at (0,0) (c2) [draw, dashed, very thick, green, rounded corners,  minimum width=1cm, minimum height=1.5cm] {};
\end{tikzpicture}
\end{center}
\caption{The figure shows a network composed of three client groups. Clients in group $j$ can sample arms in arm-set $\mc{A}_j$, where $j\in\{1,2,3\}$. Given that client $5$ is the only source of information for arms in $\mc{A}_2$, if it acts adversarially, then all other clients have no way of learning about the arms in $\mc{A}_2$. This highlights the need for information-redundancy. Moreover, since clients in $\mc{C}_1$ and $\mc{C}_3$ can interact only via client $5$, if client $5$ is adversarial, then it can create a bottleneck between $\mc{C}_1$ and $\mc{C}_3$, making it impossible for clients in $\mc{C}_1$ (resp., $\mc{C}_3$) to learn about arms in arm-set $\mc{A}_3$ (resp.,  $\mc{A}_1$). This highlights the need for redundant paths in the network. The \textit{strong-robustness} property in Definition \ref{def:strongrobustness} captures both the requirements described above.}
\label{fig:network}
\end{figure}
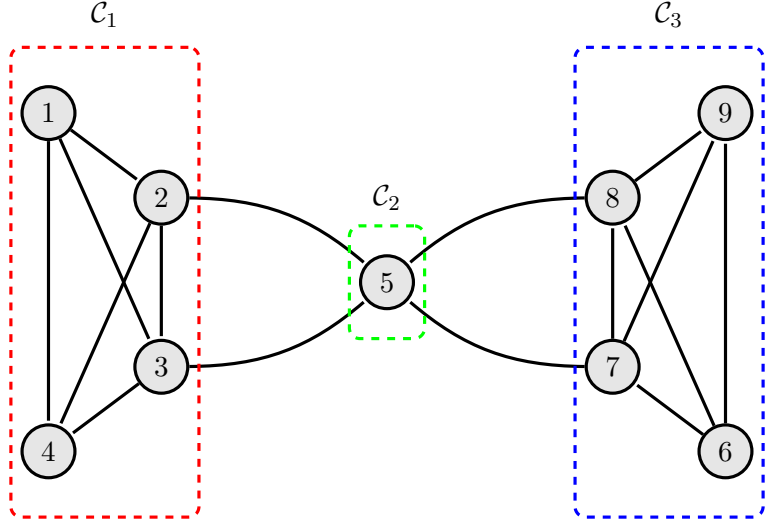

\begin{definition} 
\label{def:strongrobustness}
(\textbf{Strong-robustness w.r.t. $\mc{C}_j$}) Given any positive integer $r$, the graph $\mathcal{G}$ is said to be strongly $r$-robust w.r.t. the client group $\mathcal{C}_j$ if every non-empty subset $\mathcal{B}\subseteq\mc{V}\setminus\{\mc{C}_j\}$ contains at least one client $i$ with $r$ in-neighbors outside $\mc{B}$, i.e., for every non-empty set  $\mathcal{B}\subseteq\mc{V}\setminus\{\mc{C}_j\}$, $\exists i\in \mc{B}$ such that $|\mc{N}_i\setminus\mc{B}| \geq r$. 
\end{definition}

We note that if $\mc{G}$ is strongly $(3f+1)$-robust w.r.t. $\mathcal{C}_j$, then $|\mathcal{C}_j| \geq 3f+1$ (follows by considering $\mc{B}=\mc{V}\setminus\{\mc{C}_j\}$). Thus, the above definition implicitly captures information-redundancy. At the same time, it ensures that there is adequate redundancy in the network-structure to reliably disseminate information about arm-set $\mc{A}_j$ from $\mc{C}_j$ to the rest of the network. We now develop an algorithm, namely Algorithm \ref{algo:p2p}, that leverages the  strong-robustness property in Definition \ref{def:strongrobustness}. 

\textbf{Description of Algorithm \ref{algo:p2p}:} Each client $i$ maintains two $N$-dimensional vectors $\mathbf{I}_i$ and $\mathbf{R}_i$. The $m$-th entry of $\mathbf{I}_i$, namely $\mathbf{I}_i(m)$, stores client $i$'s estimate of the locally best arm in arm-set $\mc{A}_m$, and $\mathbf{R}_i(m)$ stores an estimate of the corresponding arm's mean. Consider a client $i\in\mathcal{C}_j\cap{\mc{H}}$. Since client $i$ can sample all arms in $\mc{A}_j$, it updates $\mathbf{I}_i(j)$ and $\mathbf{R}_i(j)$ on its own by running Algo. \ref{algo:algo1} on $\mc{A}_j$. To learn about the best arm in all arm-sets other than $\mc{A}_j$, client $i$ needs to interact with its neighbors, some of whom might be adversarial. For each $\ell \in [N]\setminus\{j\}$, the strategy employed by client $i$ to update $\mathbf{I}_i(\ell)$ and $\mathbf{R}_i(\ell)$ is very similar to that of the server in \texttt{Robust Fed-SEL}. In short, $i$ collects information about $\mc{A}_{\ell}$ from its neighbors, performs majority voting to decide $\mathbf{I}_i(\ell)$ (line 3), and does a trimming operation to compute $\mathbf{R}_i(\ell)$ (line 4). For this operation, client $i$ only uses the estimates of those neighbors that report the majority-voted arm $\mathbf{I}_i(\ell)$. Once all components of $\mathbf{I}_i$ and $\mathbf{R}_i$ have been populated, client $i$ executes line 5 to identify the globally best arm. 

We now present the main result of this section. 

\begin{theorem}
\label{thm:peertopeer}
Suppose the following arm-heterogeneity condition holds:
\begin{equation}
    \max_{j\in[N]\setminus\{1\}} \{\sigma_{1j}, \sigma_{j1}\} \leq 1.
\label{eqn:het_ass}
\end{equation}
Moreover, suppose the adversary model is $f$-local, and $\mc{G}$ is strongly $(3f+1)$-robust w.r.t. $\mc{C}_j, \forall j\in [N]$. Then, with probability at least $1-\delta$, Algorithm \ref{algo:p2p} terminates in finite-time, and outputs $\mathbf{I}_i(p^*_i)=a^*, \forall i\in\mc{H}$. 
\end{theorem}

\textbf{Discussion:} The above result is the only one we know of that provides any guarantees against adversarial attacks for MAB problems over general networks. While the strong-robustness property in Definition \ref{def:strongrobustness} has been exploited earlier in the context of resilient distributed state estimation \cite{mitraByz} and hypothesis-testing/inference \cite{mitrahyp}, Theorem \ref{thm:peertopeer} reveals that the same graph-theoretic property ends up playing a crucial role for the best-arm identification problem we consider here.\footnote{By making a connection to bootstrap percolation theory, the authors in \cite{mitraByz} show that the strong-robustness property in Definition \ref{def:strongrobustness} can be checked in polynomial time.}

To isolate the subtleties and challenges associated with tolerating adversaries over a general network, we worked under an  arm-heterogeneity assumption. This assumption effectively bypasses the need for Phase II, i.e., the sample-complexity of a client is given by the number of arm-pulls it makes during Phase I. We believe that the ideas presented in this paper can be extended in an appropriate way to study the networked setting \textit{without} making the arm-heterogeneity assumption. 

\section{Simulations}
In this section, we verify some of our theoretical findings based on a simple simulation example. Motivated by distributed target tracking applications in cooperative learning \cite{loc1,franchi,savic,dames}, we consider a target detection problem involving 9 static sensors distributed over an environment. The environment is divided into 3 sub-regions with 3 sensors and a mobile robot in each sub-region. Each mobile robot can access the sensor readings of the sensors in its sub-region, and transmit information to a central controller. The goal is to detect a static target located in one of the sub-regions. We model this setting by mapping sensors to arms, sub-regions to arm-sets, and robots to clients. Each sensor displays binary readings: sensors that are located closer to the target have a higher probability of displaying $1$. Thus, finding the ``best arm" in this setting corresponds to localizing the target, and a robot visiting a sensor corresponds to a client sampling an arm. Ideally, to minimize battery life of the robots, we would like to locate the target with a small number of visits (sample-complexity), and only a few information exchanges with the controller (communication-complexity). 

We assume the target is located in the first sub-region that corresponds to arm-set $\mc{A}_1$. Arm rewards (sensor readings) are drawn from Bernoulli distributions with means $\{0.9, 0.9-0.05\sigma, 0.1\}$, $\{0.85, 0.8, 0.3\}$, and $\{0.7, 0.6, 0.5\}$ associated with the arm-sets (sub-regions) $\mc{A}_1, \mc{A}_2$, and $\mc{A}_3$, respectively. Based on the definition of arm-heterogeneity indices in Definition \ref{defn:arm_het},  the parameter $\sigma \in [1, 15]$ captures the level of heterogeneity in the example we have described above. In particular, it is easy to verify that for each $i\in [3]$, 
 \begin{equation}
     r_{a^{(i)}_1} - r_{a^{(i)}_2} \leq \sigma|r_{a^{(i)}_1} - r_{a^{(j)}_1}|, \forall j \in [3]\setminus \{i\}.
\label{eqn:rel_armhet}
\end{equation}

\begin{figure}[t]
\begin{tabular}{c c}
\includegraphics[scale=0.375]{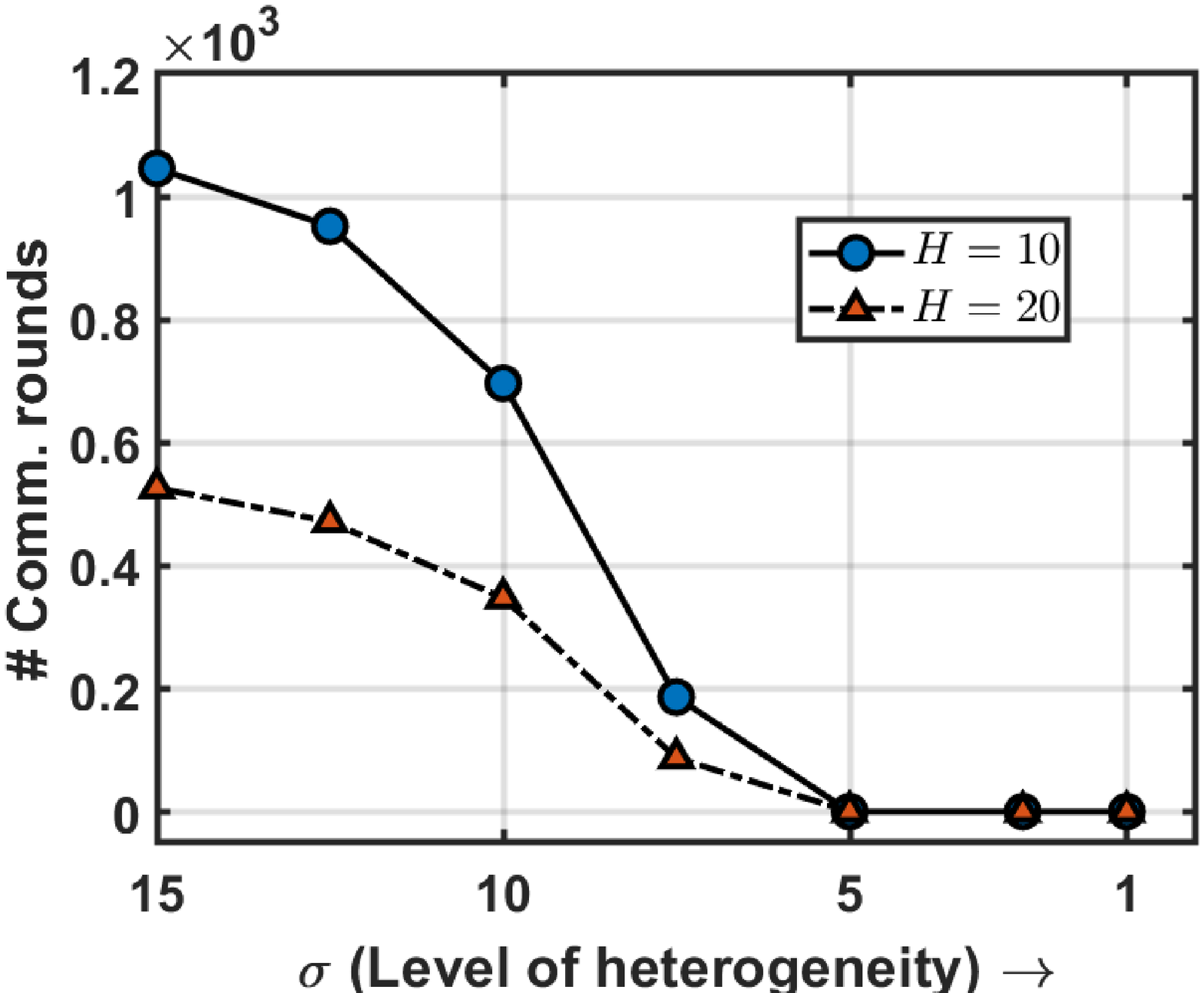}&\includegraphics[scale=0.375]{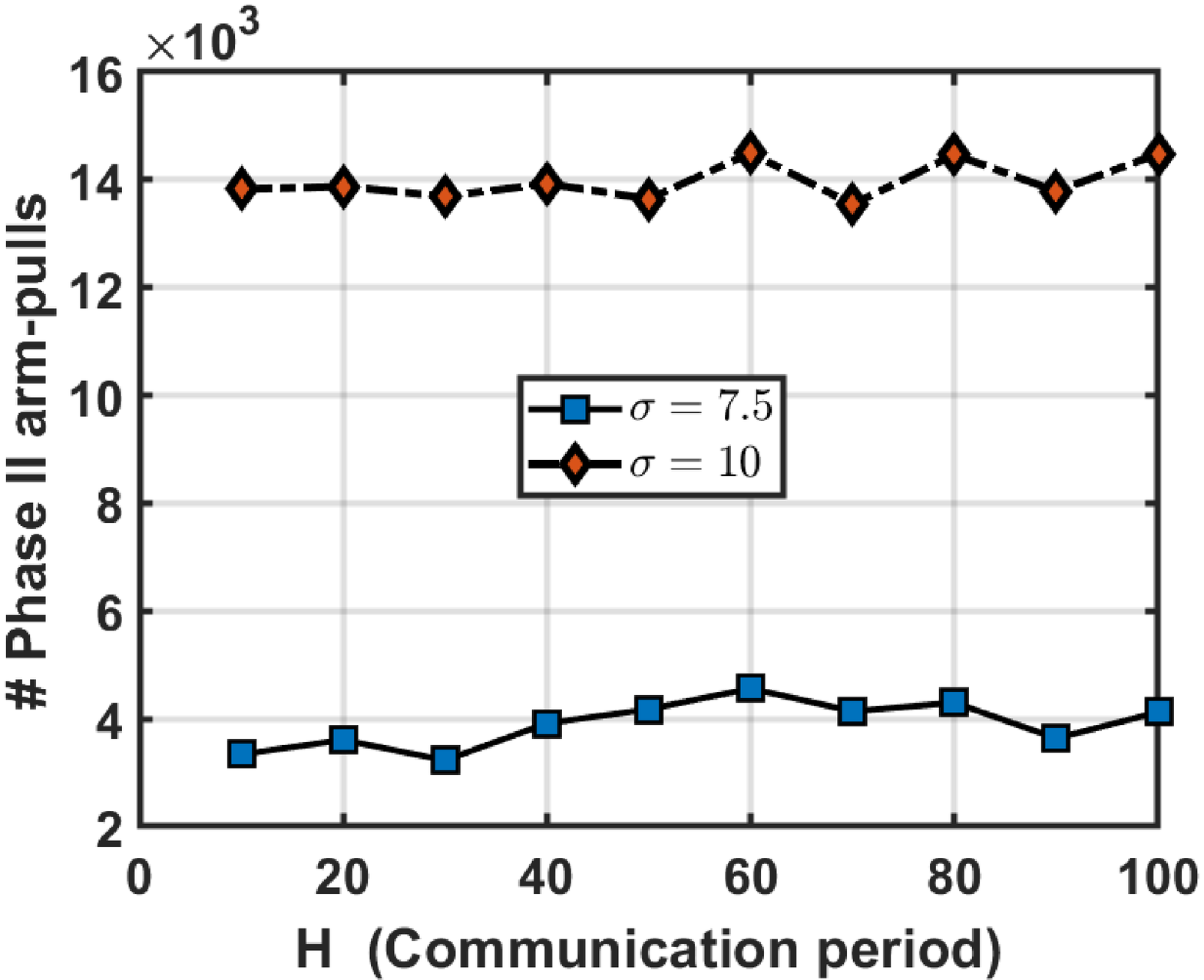}\\
(a) & (b)
\end{tabular}
\caption{Illustration of the performance of \texttt{Fed-SEL}. (a) Plot of the number of communication rounds versus the level of heterogeneity $\sigma$. (b) Plot of the number of Phase II arm-pulls versus the communication period $H$.}
\label{fig:plots}
\end{figure}

For the example we have considered, arm-heterogeneity  has a very natural spatial interpretation: \textit{sensors located closer to the target output higher readings than those farther away}. Note that arm 1 in $\mc{A}_1$ is the best arm, and hence closest to the target. With confidence parameter $\delta=0.1$, we implement 
a simplified version of \texttt{Fed-SEL} (Algo. \ref{algo:algo2}) without any adaptive quantization. We vary the level of heterogeneity by tuning $\sigma$, and for each value of $\sigma$, perform $50$ trials.

\vspace{2mm}
 \textbf{Main Observations:}  Below, we discuss our experimental findings from Figure \ref{fig:plots}. 
\begin{itemize}
    \item From Fig.~\ref{fig:plots}(a), we note a distinct trend: the number of communication rounds decrease with an increase in the level of  arm-heterogeneity $\sigma$, exactly as suggested by Theorem \ref{thm:Fed-SEL}. Moreover, when $\sigma=1$, \texttt{Fed-SEL} terminates in 1 round, complying with Corollary \ref{corr:oneround}. We conclude that \texttt{Fed-SEL} \textit{does indeed benefit from the notion of heterogeneity that we have introduced}.
    
    \item  From Fig.~\ref{fig:plots}(a), we note that doubling the communication period $H$ roughly halves the number of communication rounds, as expected. Importantly, increasing the communication period $H$ does not cause the number of arm-pulls in Phase II to increase much; see Fig.~\ref{fig:plots}(b). These observations once again align with our theoretical findings in Theorem \ref{thm:Fed-SEL}: \textit{since local computations always help for \texttt{Fed-SEL}, one can significantly reduce communication by increasing $H$, and at the same time, not pay much of an additional price in terms of sample-complexity.} 
\end{itemize}

\section{Conclusion and Future Work}
We introduced and studied the federated best-arm identification problem where clients with disjoint arm-sets collaborate via a server to identify the globally best arm with the highest mean pay-off. We defined a novel notion of arm-heterogeneity for this problem, and proved that when such heterogeneity is exploited in a principled way, one can significantly reduce the number of communication rounds required to identify the best arm (with high probability). Our finding is unique in that it reveals that unlike federated supervised learning, heterogeneity can provably help for best-arm identification problems in stochastic multi-armed bandits. Finally, a key advantage of our algorithmic approach is that it is robust to worst-case adversarial actions. 

\newpage
There are several interesting avenues of future research, as we discuss below.

\begin{itemize}
    \item First, we aim to derive lower-bounds on the sample-complexity for the problem we have introduced. This will shed light on the tightness of our results in Theorem \ref{thm:Fed-SEL}, and inform the need for potentially better algorithms. It would also be interesting to explore alternate formulations where the communication budget is fixed a priori, and the goal is to maximize the probability of identifying the best arm subject to such a budget. 
    
    \item Perhaps the most interesting direction is to see if the ideas developed in this paper have immediate extensions to general Markov decision processes (MDP's) in the context of reinforcement learning. Specifically, we wish to investigate the trade-offs between communication-complexity and performance measures such as PAC bounds or regret for distributed sequential decision-making problems. 
    
    \item Another important direction is to explore related settings where statistical heterogeneity can provably help. Our conjecture is that for problems where the globally optimal parameter belongs to a \textit{finite set}, one can establish results similar to those in this paper by using \textit{elimination-based}  techniques. In this context, one ripe candidate problem is that of hypothesis-testing/statistical inference where the true hypothesis belongs to a finite set of possible hypotheses. Recent results have shown that for peer-to-peer versions of this problem \cite{mitraCDC19,mitraTSP}, one can design principled approaches to significantly reduce communication. 
    
    \item Some recent works have shown that by either considering alternate problem formulations (such as personalized variants of FL \cite{pFL2,pFL4,pFL5,pfL6}), or by drawing on ideas from representation learning \cite{rep}, one can train meaningful statistical models in FL, despite statistical heterogeneity. It may be worthwhile to investigate whether such ideas apply to the setting we consider. Finally, while we studied robustness to adversarial actions, it would also be interesting to explore robustness to other challenges, such as distribution shifts \cite{dist_shft}, or stragglers \cite{FedNova,FLANP,FedLin}. 
\end{itemize}
\bibliographystyle{unsrt} 
\bibliography{refs}

\newpage
\appendix
\section{Proof of Theorem \ref{thm:Fed-SEL}}
\label{app:thm1proof}

The goal of this section is to prove Theorem \ref{thm:Fed-SEL}. Throughout the proof, to simplify notation, we will use $i_1$ as a shorthand to represent the best arm $a^{(i)}_1$ in arm-set $\mc{A}_i$. Also, for any arm $\ell \in \mc{A}_i$, we will use $T_{\ell}$ to denote the number of pulls made to arm $\ell$ by client $i$ during the course of operation of \texttt{Fed-SEL}. Before diving into the formal argument, we first outline the main steps of the proof. 

\begin{itemize}
    \item \textbf{Step 1:} The first step of our analysis is to define a ``good event" $\mc{E}$, and show that it occurs with probability at least $1-\delta$, where $\delta \in (0,1)$ is the given confidence parameter. On this good event, the empirical estimates of all arms are within appropriate confidence intervals (as given by the $\alpha's$) of the true means. See Lemma \ref{lemma:good_event}. 
    
    \item \textbf{Step 2:}  On $\mc{E}$, we show that the local subroutine (Algo. \ref{algo:algo1}) at each client $i\in\mc{C}$ terminates in finite-time, and outputs the locally best-arm $a^{(i)}_1$ in $\mc{A}_i$, i.e., $a_i=a^{(i)}_1, \forall i\in\mc{C}$. This step follows from standard arguments \cite{even2002,even}; we provide a proof only for completeness. See Lemma \ref{lemma:Algo1}. 
    
    \item \textbf{Step 3:} The next key step of our analysis is to bound the error of each client $i$'s estimate $\hat{r}_{i,a_i}(\bar{t}_i)$ of the true mean ${r}_{i_1}$, at the end of Phase I, as a function of the local arm-gap   $\Delta^*_i=r_{i_1}-r_{i_2}$. This analysis, carried out in Lemma \ref{lemma:alpha_bnd}, reveals that the locally best arm from each arm-set is ``well-explored" at the end of Phase I. 
    
    \item \textbf{Step 4:} The final piece of the analysis is to relate the amount of exploration that has been done during Phase I, to that which remains to be done during Phase II, for eliminating each client in the set $\mc{C}\setminus\{1\}.$ Specifically, for each client $i\in \mc{C}\setminus\{1\}$, we seek a lower bound on the number of times arm $i_1$ has been sampled during Phase I, and an upper bound on the number of times it still needs to be sampled during Phase II, before being deemed sub-optimal. See Lemma \ref{lemma:time_diff_bnd}. To complete this step, we derive new bounds on the Lambert $W$ function in Lemma \ref{lemma:lambert} of Appendix \ref{app:lambert_bnds}. These bounds may be of independent interest. 
\end{itemize}

We now proceed to establish each of the above steps, starting with Step 1. 

\begin{lemma}
\label{lemma:good_event}
Consider the event $\mc{E}$ defined as follows:
\begin{equation}
    \mc{E} \triangleq \bigcap_{i\in\mc{C}} \bigcap_{\ell \in\mc{A}_i} \bigcap_{t\in [T_\ell]} \{ \vert \hat{r}_{i,\ell}(t) - r_{\ell} \vert \leq \alpha_i(t) \}. 
\label{eqn:event}
\end{equation}
Then, it holds that $\mathbb{P}(\mc{E}) \geq 1-\delta$. 
\end{lemma}
\begin{proof}
We can assume that the sequence of rewards for each arm is drawn before the algorithm is initiated. That way, for any $t\in \{1,2,\ldots, \infty\}$, the empirical estimate of the mean of an arm $\ell$ after $t$ pulls is well defined, even if it has not actually been pulled $t$ times by a client. Applying the union bound gives us
\begin{equation}
    \begin{aligned}
    \mathbb{P}(\mc{E}^{c}) &\leq \sum_{i\in\mc{C}} \sum_{\ell \in \mc{A}_i} \sum\limits_{t=1}^{\infty} \mathbb{P} \left( \{\vert \hat{r}_{i,\ell}(t) - r_{\ell} \vert > \alpha_i(t) \} \right) \\
    & \leq \sum_{i\in\mc{C}} \sum_{\ell \in \mc{A}_i} \sum\limits_{t=1}^{\infty} 2 \exp(-2 t \alpha^2_i(t)) \\
    & \leq \sum_{i\in\mc{C}} \sum_{\ell \in \mc{A}_i} \sum\limits_{t=1}^{\infty} \frac{\delta}{2 |\mc{C}| |\mc{A}_i| t^2}\\
    & \leq \sum_{i\in\mc{C}} \sum_{\ell \in \mc{A}_i}  \frac{\delta}{2 |\mc{C}| |\mc{A}_i|} \left(1+\int_{1}^{\infty} \frac{1}{z^2} dz\right) \\
    & = \sum_{i\in\mc{C}} \sum_{\ell \in \mc{A}_i}  \frac{\delta}{ |\mc{C}| |\mc{A}_i|} = \delta.
    \end{aligned}
\end{equation}
The second step above follows from an application of Hoeffding's inequality \cite{hoeff}. 
\end{proof}

For Step 2, we have the following lemma. 

\begin{lemma} \label{lemma:Algo1} Suppose \texttt{Fed-SEL} is run with $\beta_i(t)= c\alpha_i(t)$, where $c \geq 2$. Then, on the event $\mc{E}$, for each client $i\in\mc{C}$, Algo. \ref{algo:algo1} terminates in finite-time and outputs $a_i = a^{(i)}_1$. 
\end{lemma}
\begin{proof}
We prove the result for the case when $c=2$; the same reasoning applies when $c >2$. Fix a client $i\in\mc{C}$. We first argue that on $\mc{E}$, the locally best arm $a^{(i)}_1$ in $\mc{A}_i$ never gets eliminated when Algo. \ref{algo:algo1} is run. To see this, observe that on $\mc{E}$, the following holds in each epoch $t$:
$$ \hat{r}_{i,i_1}(t) \geq r_{i_1} - \alpha_i(t).$$
At the same time, for any other arm $\ell \in \mc{S}_i(t) \setminus \{i_1\}$, the following also holds on $\mc{E}$:
$$ \hat{r}_{i,\ell}(t) \leq r_{\ell} + \alpha_i(t) < r_{i_1} + \alpha_i(t). $$
Thus, on $\mc{E}$, we have that
$$ \hat{r}_{i,\ell}(t) - \hat{r}_{i,i_1}(t) < 2 \alpha_i(t) = \beta_i(t). $$
From line 5 of Algo. \ref{algo:algo1}, it is then clear that $a^{(i)}_1$ does not get eliminated on $\mc{E}$. We now argue that every arm $\ell \in \mc{A}_i \setminus \{i_1\}$ eventually gets eliminated. In particular, we claim that 
$$ \alpha_i(t) \leq \frac{\Delta_{i_1\ell}}{4} $$
 is a sufficient condition to eliminate arm $\ell$. To see this, suppose the above condition holds. We then have
 \begin{equation}
 \begin{aligned}
 & r_{i_1} - \alpha_i(t) \geq r_{\ell} + \alpha_i(t) +2 \alpha_i(t) \\
\overset{\mc{E}} \implies & \hat{r}_{i,i_1} (t)  \geq \hat{r}_{i,\ell}(t) + 2\alpha_i(t) =  \hat{r}_{i,\ell}(t) + \beta_i(t) \\
 \implies & \hat{r}_{i,\max} (t) \geq \hat{r}_{i,\ell}(t) + \beta_i(t).
 \end{aligned}
 \end{equation}
 The claim then follows from line 5 of Algo. \ref{algo:algo1}. This completes the proof.
\end{proof}

The next lemma provides a bound on the confidence intervals at the end of Phase I. 

\begin{lemma} \label{lemma:alpha_bnd} Suppose \texttt{Fed-SEL} is run with $\beta_i(t)= c\alpha_i(t)$, where $c > 2$. Then, on the event $\mc{E}$, the following holds for every client $i\in\mc{C}$: 
\begin{equation}
    \alpha_i(\bar{t}_i) < \frac{c}{(c-2)^2} \Delta^*_{i},
\end{equation}
where $\Delta^*_{i}=r_{i_1}-r_{i_2}$.
\end{lemma}
\begin{proof}
Fix a client $i\in\mc{C}$. From Lemma \ref{lemma:Algo1}, we know that every arm $\ell \in \mc{A}_i\setminus\{i_1\}$ eventually gets eliminated when client $i$ runs Algo. \ref{algo:algo1} on $\mc{A}_i$. We split our subsequent analysis into two cases depending on \textit{when} the second best arm $i_2$ in $\mc{A}_i$ gets eliminated during Phase I. 

\textbf{Case 1:} Suppose arm $i_2$ is eliminated in the last epoch $\bar{t}_i$. On event $\mc{E}$, the arm with the largest empirical mean in $\mc{S}_i(\bar{t}_i)$ must be $i_1$ (from Lemma \ref{lemma:Algo1}). Thus, we must have
\begin{equation}
    \begin{aligned}
    &\hat{r}_{i,i_1}(\bar{t}_i) \geq \hat{r}_{i,i_2}(\bar{t}_i) + c \alpha_i(\bar{t}_i) \\
   \overset{\mc{E}} \implies & r_{i_1} + \alpha_i(\bar{t}_i) \geq \hat{r}_{i,i_2}(\bar{t}_i) + c \alpha_i(\bar{t}_i) \\
   \overset{\mc{E}} \implies & r_{i_1} + \alpha_i(\bar{t}_i) \geq r_{i_2} - \alpha_i(\bar{t}_i) + c \alpha_i(\bar{t}_i) \\
\implies & \alpha_i(\bar{t}_i) \leq \frac{\Delta^*_i}{(c-2)}.
    \label{eqn:alphabnd1}
    \end{aligned}
\end{equation}

\textbf{Case 2:} Suppose arm $i_2$ is eliminated in some epoch $t < \bar{t}_i$. For this to happen on $\mc{E}$, it must be that $i_1 = \argmax_{\ell \in \mc{S}_i(t)} \hat{r}_{i,\ell}(t)$. This is easy to verify. Moreover, note that there must exist some arm $\ell \in \mc{A}_i \setminus \{i_1 \cup i_2 \}$ that does not get eliminated in epoch $t$, but gets eliminated in epoch $\bar{t}_i$. Thus, the following inequalities must hold simultaneously:
\begin{equation}
    \hat{r}_{i,i_1}(t) - \hat{r}_{i,i_2}(t) \geq c \alpha_i(t); \hspace{1mm} \hat{r}_{i,i_1}(t) - \hat{r}_{i,\ell}(t) < c \alpha_i(t),
\end{equation}
implying $\hat{r}_{i,\ell}(t) > \hat{r}_{i,i_2}(t)$. Since we are on $\mc{E}$, this further implies that
\begin{equation}
    r_{\ell} + \alpha_i(t) > r_{i_2} - \alpha_i(t) \implies 2 \alpha_i (t) > \Delta_{i_2 \ell}.
\label{eqn:alphabnd2}
\end{equation}
At the same time, since arm $i_2$ gets eliminated in epoch $t$, we must also have 
\begin{equation}
\alpha_i(t) \leq \frac{\Delta^*_i}{(c-2)},
\end{equation}
following the analysis of Case 1. Combining Eq. \eqref{eqn:alphabnd2} with the above bound yields
\begin{equation}
\begin{aligned}
& \Delta_{i_2 \ell} < 2 \frac{\Delta^*_i}{(c-2)}\\
\implies & r_{i_2} - r_{i_1} +r_{i_1} - r_{\ell} < 2 \frac{\Delta^*_i}{(c-2)} \\
\implies & \Delta_{i_1 \ell} < \frac{c}{(c-2)} \Delta^*_i.
\end{aligned}
\label{eqn:alphabnd3}
\end{equation}
Finally, since arm $\ell$ gets eliminated in the last epoch $\bar{t}_i$, an argument identical to the Case 1 analysis reveals that 
\begin{equation}
\alpha_i(\bar{t}_i) \leq  \frac{\Delta_{i_1\ell}}{(c-2)}. 
\end{equation}
Combining the above inequality with the one in Eq. \eqref{eqn:alphabnd3}, we obtain 
\begin{equation}
    \alpha_i(\bar{t}_i) < \frac{c}{(c-2)^2} \Delta^*_{i}.
\label{eqn:alphabnd4}
\end{equation}
This completes the analysis for Case 2. The claim of the lemma then follows immediately from equations \eqref{eqn:alphabnd1} and \eqref{eqn:alphabnd4}. 
\end{proof}

Before proceeding to our next result, we need to first take a brief detour and familiarize ourselves with the Lambert $W$ function that will show up in our subsequent analysis. Accordingly, we start by noting that the Lambert $W$ function is defined to be the multi-valued inverse of the function $f(w)=we^w$, where $w\in\mathbb{C}$ is any complex number \cite{lambertw}.\footnote{Here, and in what follows, we use $e$ to denote the exponential function.} Specifically, the Lambert $W$ function, denoted by $W(x)$, has the following defining property:
\begin{equation}
    W(x) e^{W(x)} =x, \hspace{1mm} \textrm{where} \hspace{1mm} x\in \mathbb{C}.
\label{eqn:Lambert_def}
\end{equation}
When $x$ is real and belongs to $[-1/e, 0)$, $W(x)$ can take on two possible values. The branch satisfying $W(x) \geq -1$ is called the \textit{principal branch} and is denoted by $W_0(x)$. The other branch with values satisfying $W(x) \leq -1$ is denoted by $W_{-1}(x)$. The two branches meet at the point $x=-1/e$, where they both take on the value of $-1$. For a depiction of these branches, see Fig. \ref{fig:lamb_bounds}. 

We now state and prove the final ingredient required for establishing Theorem \ref{thm:Fed-SEL}. 

\begin{lemma}
\label{lemma:time_diff_bnd}
For $t\in[1,\infty)$, define $\alpha(t) \triangleq \sqrt{2\log(\bar{c}t)/t}$, where $\bar{c} > e$. Given $\Delta \in (0,1)$ and $\sigma >1$ such that $\sigma \Delta \leq 1$, suppose $\tilde{t}$ and $t'$ satisfy:
$$ \alpha(\tilde{t}) =\frac{\sigma \Delta}{8}, \hspace{2mm} \textrm{and} \hspace{2mm} \alpha(t') =\frac{\Delta}{8}.$$ Then, we have
\begin{equation}
    t'-\tilde{t} \leq \left( \frac{128}{\Delta^2} \log\frac{128\bar{c}}{\Delta^2} - \frac{128}{\sigma^2 \Delta^2} \log\frac{128\bar{c}}{\sigma^2\Delta^2} \right) + \left( \frac{512}{\Delta^2} \log\log\frac{128\bar{c}}{\Delta^2} - \frac{32}{\sigma^2 \Delta^2} \log\log\frac{128\bar{c}}{\sigma^2\Delta^2} \right).
\label{eqn:time_diff}
\end{equation}
\end{lemma}

\begin{proof}
Our goal is to leverage the bounds on the $W_{-1}(x)$ branch that we derived in Lemma \ref{lemma:lambert}. To do so, we need to first transform the equations concerning $\alpha(\tilde{t})$ and $\alpha(t')$ into an appropriate form. Accordingly, simple manipulations yield:
\begin{equation}
    \left(-\frac{\sigma^2 \Delta^2 \tilde{t}}{128}\right) \exp{\left(-\frac{\sigma^2 \Delta^2 \tilde{t}}{128}\right)}= -\frac{\sigma^2 \Delta^2 }{128 \bar{c}}; \hspace{2.5mm} \left(-\frac{\Delta^2 {t}'}{128}\right) \exp{\left(-\frac{\Delta^2 {t}'}{128}\right)}= -\frac{ \Delta^2 }{128 \bar{c}}. 
\end{equation}
Based on the conditions of the lemma, it is easy to see that both $-\frac{\sigma^2 \Delta^2 }{128 \bar{c}}$ and $-\frac{ \Delta^2 }{128 \bar{c}}$ belong to the interval $[-1/e^2,0).$ Using the defining property of the Lambert $W$ function in Eq. \eqref{eqn:Lambert_def}, we claim:
\begin{equation}
    \tilde{t}= -\frac{128}{\sigma^2 \Delta^2} W_{-1}\left(-\frac{\sigma^2 \Delta^2 }{128 \bar{c}}\right); \hspace{2mm} {t}'= -\frac{128}{\Delta^2} W_{-1}\left(-\frac{\Delta^2}{128 \bar{c}}\right).
\label{eqn:solutions_lamb}
\end{equation}
We need to now argue why the above solutions for $\tilde{t}$ and $t'$ involve the $W_{-1}(x)$ branch, and not the principal branch $W_0(x)$. To see why, note that $W_0(x) \geq ex, \forall x\in [-1/e^2,0)$. This in turn follows by observing that:
$$ W_0(x) \geq -1 \implies e \geq e^{-W_0(x)}  \implies ex \leq x  e^{-W_0(x)} = W_0(x), $$
where we used $x < 0$ and Eq. \eqref{eqn:Lambert_def}. Now suppose the solution for $\tilde{t}$ in Eq. \eqref{eqn:solutions_lamb} involves $W_0(x)$ instead of $W_{-1}(x)$. Using $W_0(x) \geq ex$ then leads to the following:
$$ \tilde{t}= -\frac{128}{\sigma^2 \Delta^2} W_{0}\left(-\frac{\sigma^2 \Delta^2 }{128 \bar{c}}\right) \leq \frac{e}{\bar{c}} < 1,$$
since $\bar{c} > e$. This leads to a contradiction as $\alpha(t)$ is only defined for $t\in[1,\infty)$. A similar argument can be made for the solution involving $t'$. We have thus justified \eqref{eqn:solutions_lamb}. 

We can now directly appeal to the bounds in Lemma \ref{lemma:lambert} to obtain
    \begin{equation}
    W_{-1}\left(-\frac{\sigma^2 \Delta^2 }{128 \bar{c}}\right) \leq \log\left(\frac{\sigma^2 \Delta^2}{128\bar{c}}\right) -\frac{1}{4}\log \log \frac{128\bar{c}}{\sigma^2 \Delta^2}; \hspace{2mm}  W_{-1}\left(-\frac{ \Delta^2 }{128 \bar{c}}\right) \geq \log\left(\frac{ \Delta^2}{128\bar{c}}\right) -4\log \log \frac{128\bar{c}}{\Delta^2}. 
    \end{equation}
Using the above bounds in conjunction with the expressions for $\tilde{t}$ and $t'$ in Eq. \eqref{eqn:solutions_lamb} immediately leads to the claim of the lemma. 
\end{proof}

We are now in position to complete the proof of Theorem \ref{thm:Fed-SEL}. 
\begin{proof} (\textbf{Theorem \ref{thm:Fed-SEL}}) We first argue that \texttt{Fed-SEL} is consistent, i.e., it outputs the globally best arm. To this end, we claim that on the event $\mc{E}$ in Eq. \eqref{eqn:event}, client $1$ always remains active during Phase II (recall that $a^* = 1 \in \mc{A}_1$). To see why this is true, fix any client $i\in \mc{C}$, and note that based on our encoding-decoding strategy, 
\begin{equation}
\begin{aligned}
    \vert \tilde{r}_i(k) - \hat{r}_{i,a_i}(\bar{t}_i +kH) \vert & \leq  \frac{1}{2} \frac{1}{2^{B_i(k)}}\\
    &\leq \frac{\alpha_i(\bar{t}_i+kH)}{2}, \forall k \geq 1,
\label{eqn:quant_err}
\end{aligned}
\end{equation}
where we used the expression for $B_i(k)$ in Eq. \eqref{eqn:bit_prec}. The above bound holds trivially when $k=0$, since $\tilde{r}_i(0)=\hat{r}_{i,a_i}(\bar{t}_i)$. Next, observe that at any round $k$, 
\begin{equation}
\begin{aligned}
    \tilde{r}^{(L)}_{i}(k) &=\tilde{r}_i(k) - 2 \alpha_i(\bar{t}_i+kH) \\ 
    & \overset{\eqref{eqn:quant_err}} \leq \hat{r}_{i,a_i}(\bar{t}_i +kH) + \frac{\alpha_i(\bar{t}_i+kH)}{2} - 2 \alpha_i(\bar{t}_i+kH) \\
    & \overset{(a)}= \hat{r}_{i,i_1}(\bar{t}_i+kH) - \frac{3}{2}\alpha_i(\bar{t}_i+kH)\\
    &\overset{\mc{E}}\leq {r_{i_1}} + \alpha_i(\bar{t}_i+kH)-\frac{3}{2}\alpha_i(\bar{t}_i+kH)\\
    & < r_{i_1}.
\label{eqn:FSEL_lb1}
\end{aligned}
\end{equation}
Here, (a) follows from Lemma \ref{lemma:Algo1} where we showed that on $\mc{E}$, ${a}_i = a^{(i)}_1=i_1, \forall i\in\mc{C}$. Using the same reasoning as above, we can show that
\begin{equation}
\tilde{r}^{(U)}_{i}(k) > r_{i_1}, \forall i\in \mc{C}.
\label{eqn:FSEL_ub1}
\end{equation}
From \eqref{eqn:FSEL_lb1} and \eqref{eqn:FSEL_ub1}, we conclude that on $\mc{E}$, 
$$ \tilde{r}^{(U)}_{1}(k) > r_1 \hspace{1mm} \textrm{and} \hspace{1mm}  \tilde{r}^{(L)}_{i}(k) < r_{i_1}, \forall i\in \mc{C}\setminus \{1\}.$$
Based on lines 4 and 5 of \texttt{Fed-SEL}, it is easy to then see that client $1$ always remains in the active-client set $\bar{\Psi}(k)$ on the event $\mc{E}$. 

We now show that every client $i\in\mc{C}\setminus\{1\}$ eventually gets removed from the active-client set. In particular, for a client $i\in\mc{C}\setminus\{1\}$ to get removed, we claim that the following is a set of sufficient conditions: 
\begin{equation}
    \begin{aligned}
    \alpha_i(\bar{t}_i+kH) &\leq \frac{\Delta_{1i_1}}{8},  \textrm{and} \\
    \alpha_1(\bar{t}_1+kH) &\leq \frac{\Delta_{1i_1}}{8}.
    \end{aligned}
\label{eqn:suff_cond}
\end{equation}
Suppose both the above conditions hold. We then have
\begin{equation}
\begin{aligned}
    \tilde{r}^{(U)}_{i}(k) &=\tilde{r}_i(k) + 2 \alpha_i(\bar{t}_i+kH) \\ 
    & \overset{\eqref{eqn:quant_err}} \leq \hat{r}_{i,a_i}(\bar{t}_i +kH) + \frac{\alpha_i(\bar{t}_i+kH)}{2} + 2 \alpha_i(\bar{t}_i+kH) \\
    & \overset{\mc{E}}= \hat{r}_{i,i_1}(\bar{t}_i+kH) +  \frac{5}{2}\alpha_i(\bar{t}_i+kH)\\
    &\overset{\mc{E}}\leq {r_{i_1}} + \alpha_i(\bar{t}_i+kH)+\frac{5}{2}\alpha_i(\bar{t}_i+kH)\\
    & \overset{\eqref{eqn:suff_cond}} \leq  r_{i_1} + \frac{7}{16} \Delta_{1i_1}.
\label{eqn:FSEL_ub2}
\end{aligned}
\end{equation}
Similarly, we have
\begin{equation}
    \tilde{r}^{(L)}_{1}(k) \geq r_{1} - \frac{7}{16} {\Delta_{1i_1}}.
\label{eqn:FSEL_lb2}
\end{equation}

From \eqref{eqn:FSEL_ub2} and \eqref{eqn:FSEL_lb2}, we note that $\tilde{r}^{(L)}_{1}(k) >  \tilde{r}^{(U)}_{i}(k)$, which is sufficient for client $i$ to get removed from the active-client set (based on line 5 of \texttt{Fed-SEL}). Thus, we have argued that on $\mc{E}$, \texttt{Fed-SEL} will eventually terminate with $\bar{\Psi}(k+1)=\{1\}$, for some round $k$. The claim that \texttt{Fed-SEL} is $(0,\delta)$-PAC then follows immediately by noting that on $\mc{E}$, $a_1=a^*=1$. 

We now derive bounds on the communication-complexity of \texttt{Fed-SEL}. For any $i\in\mc{C}\setminus\{1\}$, let $R_{i1}$ (resp., $R_{1i}$) be the first round such that $\alpha_i(\bar{t}_i+kH) \leq \Delta_{1i_1}/{8}$ (resp., $\alpha_1(\bar{t}_1+kH) \leq \Delta_{1i_1}/{8}$). Based on the fact that $\alpha_i(t)$ is monotonically decreasing for all $i\in\mc{C}$ (this is easy to verify), and the above analysis, client $i$ remains active for at most $R_i =\max\{R_{i1},R_{1i}\}$ rounds. To bound $R_{i1}$, we start by noting that with $c=8$ in Lemma \ref{lemma:alpha_bnd}, 
$$ \alpha_i(\bar{t}_i) < \frac{2}{9} \Delta^*_i <  \frac{\Delta^*_i}{8} = \frac{\sigma_{i1} \Delta_{1i_1}}{8}, $$
where for the last step we used the definition of the arm-heterogeneity index $\sigma_{i1}$ in Eq. \eqref{eqn:arm_het}. Clearly, if $\sigma_{i1} \leq 1$, then $R_{i1}=1$. Now suppose $\sigma_{i1} >1$. In this case, we have
\begin{equation}
    R_{i1} \leq \ceil*{\frac{1}{H}\left( \ceil*{\alpha^{-1}_i\left(\frac{\Delta_{1i_1}}{8}\right)} - \floor*{\alpha^{-1}_i\left(\frac{\sigma_{i1}\Delta_{1i_1}}{8}\right)}\right)}.
\label{eqn:Ri1bound1}
\end{equation}
We now apply Lemma \ref{lemma:time_diff_bnd} with
$$ \alpha(t)=\alpha_i(t), \Delta= \Delta_{1i_1}, \sigma=\sigma_{i1}, \hspace{1mm} \textrm{and} \hspace{1mm} \bar{c}= \bar{c}_i = \sqrt{\frac{4|\mc{C}| |\mc{A}_i|}{\delta}}.$$ 
Since $|\mc{A}_i| \geq 2$, we have $\bar{c}_i > e$. Moreover, $\sigma_{i1}\Delta_{1i_i}=\Delta^*_i \leq 1$. Thus, Lemma \ref{lemma:time_diff_bnd} is indeed applicable. From \eqref{eqn:time_diff} and \eqref{eqn:Ri1bound1}, we then have 
$$ R_{i1} \leq \frac{1}{H} \left( \underbrace{ \frac{128}{\Delta^2_{1i_1}} \log\frac{128\bar{c}_i}{\Delta^2_{1i_1}} - \frac{128}{\sigma^2_{i1} \Delta^2_{1i_1}} \log\frac{128\bar{c}_i}{\sigma^2_{i1} \Delta^2_{1i_1}}}_{T_1}+T_2  \right) + 1,$$ 
where 
$$ T_2 = \frac{512}{\Delta^2_{1i_1}} \log\log\frac{128\bar{c}_i}{\Delta^2_{1i_1}} - \frac{32}{\sigma^2_{i1} \Delta^2_{1i_1}} \log\log\frac{128\bar{c}_i}{\sigma^2_{i1} \Delta^2_{1i_1} }+2 = O\left(\frac{1}{\Delta^2_{1i_1}} \log\log\frac{\bar{c}_i}{\Delta^2_{1i_1}}\right).$$ 
Using $\sigma_{i1}\Delta_{1i_1}=\Delta^*_i$ to simplify the expression for $T_1$, we obtain
$$ T_1= \frac{128}{{(\Delta^*_i)}^2} \left(\sigma^2_{i1}-1\right)  \log\frac{128\bar{c}_i}{\Delta^2_{1i_1}} +  \frac{256}{{(\Delta^*_i)}^2} \log{\sigma_{i1}}. $$ 
Combining all the above pieces together yields the upper bound on $R_{i1}$ in Eq. \eqref{eqn:Ri1}. Following exactly the same reasoning as above, one can derive the upper bound on  $R_{1i}$ in Eq. \eqref{eqn:R1i}. 

The total number of communication rounds $R$ of \texttt{Fed-SEL} equals the number of rounds it takes to eliminate every client in the set $\mc{C}\setminus\{1\}$. Thus, $R=\max_{i\in\mc{C}\setminus\{1\}} R_i.$ Regarding the claim about the number of bits exchanged, note from the definition of $R_{i1}$ that for every round $k\leq R_{i1}$, 

$$ \alpha_i(\bar{t}_i+kH) \geq \frac{\Delta_{1i_1}}{8} \geq \frac{\Delta^*}{8},$$

where we used the definition of $\Delta^*.$ From the expression for $B_i(k)$ in Eq. \eqref{eqn:bit_prec}, we then have

$$ B_i(k) \leq \log_2\left({\frac{1}{\alpha_i(\bar{t}_i+kH)}}\right)+1 \leq \log_2\left(\frac{8}{\Delta^*}\right)+1 = O\left(\log_2\left(\frac{1}{\Delta^*}\right)\right).$$

This establishes the communication-complexity of \texttt{Fed-SEL}. 

Finally, we derive the sample-complexity bounds for \texttt{Fed-SEL} in Eq. \eqref{eqn:arm_pulls}. For any client $i$, the total number of arm-pulls it makes is the sum of the number of pulls during Phases I and II. During Phase II, client $i$ only samples arm $a_i=i_1$ from its arm-set on the event $\mc{E}$. Moreover, it keeps sampling $i_1$ for the entire duration it remains in the active-client set. Since $i_1$ is sampled $H$ times in each round, and client $i$ remains active for at most $R_i$ rounds, the number of arm-pulls made by client $i$ during Phase II is $HR_i$. Since client 1 remains active till the termination of \texttt{Fed-SEL}, we have $R_1=R$.

For Phase I, recall from Lemma \ref{lemma:Algo1} that an arm $\ell \in \mc{A}_i \setminus \{i_1\}$ gets removed from the set $\mc{S}_i(t)$ when 
$$
    \alpha_i(t) \leq \frac{\Delta_{i_1 \ell}}{4},
$$
which holds for 
$$ t=O\left( \frac{\log\left(\frac{|\mc{C}| |\mc{A}_i|}{\delta \Delta_{i_1 \ell}}\right)}{\Delta^2_{i_1 \ell}} \right).$$

The above discussion immediately leads to the sample-complexity bound in Eq. \eqref{eqn:arm_pulls}. This completes the proof of Theorem \ref{thm:Fed-SEL}. 
\end{proof}
\newpage
\subsection{Bounds on the Lambert W function}
\label{app:lambert_bnds}
\begin{figure}[t]
\centering
\begin{tabular}{cc}
\includegraphics[scale=0.475]{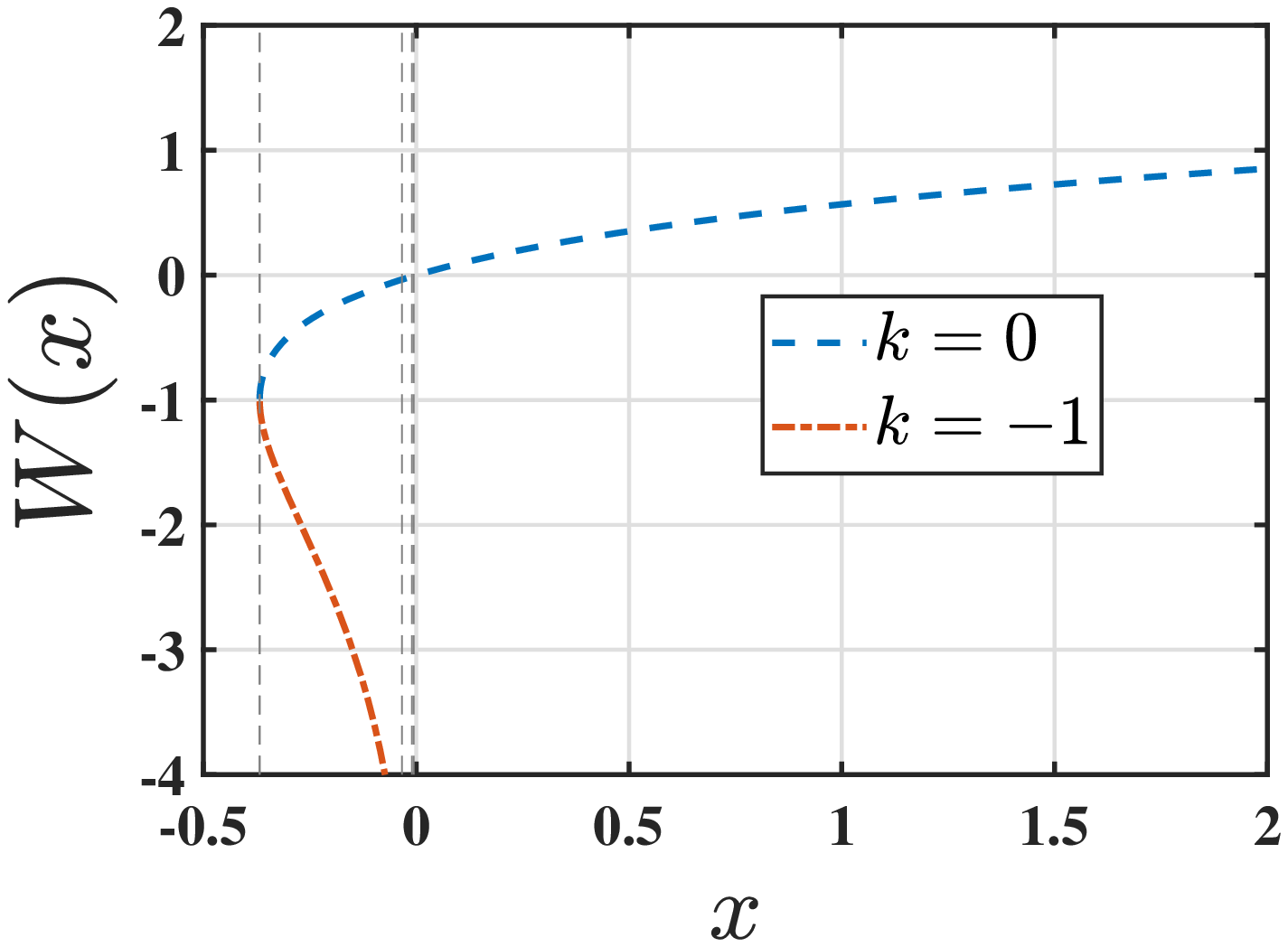}&\hspace{-3mm}\includegraphics[scale=0.475]{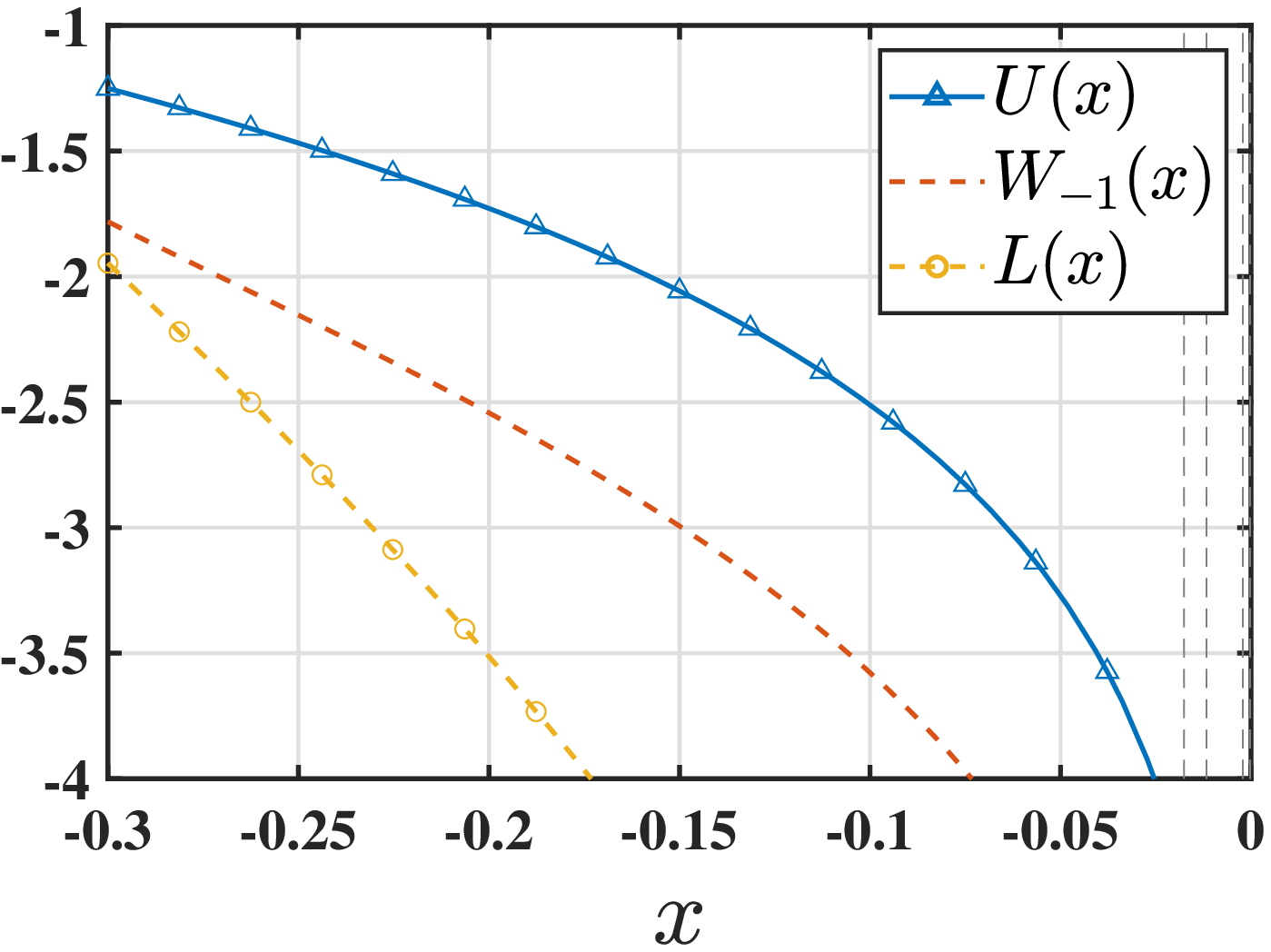}
\end{tabular}
\caption{(\textbf{Left}) Illustration of the two branches of the Lambert $W$ function. (\textbf{Right}) Illustration of the bounds in Lemma \ref{lemma:lambert}. In this figure, $U(x)$ and $L(x)$ are the upper and lower bounds, respectively, on $W_{-1}(x)$ that we derive in Lemma \ref{lemma:lambert}; see Eq. \eqref{eqn:lambert_bnd}.} 
\label{fig:lamb_bounds}
\end{figure}
In this section, we will derive new bounds on the Lambert $W$ function. To derive our bounds, we will rely on the following two facts: (i) $W_{-1}(x) < -1, \forall x\in [-1/e^2, 0)$, and (ii) $ -4 < W_{-1}(-1/e^2) < -3$. The first fact follows readily by noting that $W_{-1}(-1/e) = -1$, and that $W_{-1}(x)$ is a strictly decreasing function on $(-1/e, 0)$; see \cite{lambertw}. For the second fact, one can readily use any standard mathematical programming software (such as MATLAB) to verify that $W_{-1}(-1/e^2) \approx -3.1462.$ Alternatively, one can use the bounds derived in \cite{lambert_2} to arrive at the second fact. We have the following result. For a pictorial depiction of our bounds, see Fig. \ref{fig:lamb_bounds}.

\begin{lemma}
\label{lemma:lambert}
For $x\in[-1/e^2,0)$, we have 
\begin{equation}
    \log(-x)-4\log(-\log(-x)) \leq W_{-1}(x) \leq \log(-x)-\frac{1}{4}\log(-\log(-x)).
\label{eqn:lambert_bnd}
\end{equation}
\end{lemma}
\begin{proof}
We start by defining a function $f_{\gamma}(x)$ parameterized by $\gamma$:
\begin{equation}
    f_{\gamma}(x)=\log(-x)-\gamma \log\left(-\log(-x)\right) - W_{-1}(x).
\label{eqn:f_defn}
\end{equation}
Using $f'_{\gamma}(x)$ to denote the derivative of $f_{\gamma}(x)$ w.r.t. $x$, we obtain
\begin{equation}
\begin{aligned}
    f'_{\gamma}(x) &= \frac{1}{x}\left[1-\frac{\gamma}{\log(-x)} - \frac{W_{-1}(x)}{1+W_{-1}(x)} \right] \\
    &= \frac{1}{x} \left[\frac{\log(-x)-\gamma \left(1+W_{-1}(x)\right)}{\log(-x)\left(1+W_{-1}(x)\right)} \right].
\end{aligned}  
\label{eqn:f_derive}
\end{equation}
Here, we used the fact that for $x\in[-1/e^2,0)$, the derivative of $W_{-1}(x)$ is given by \cite{lambertw} 
$$ \frac{W_{-1}(x)}{x(1+W_{-1}(x))}.$$
Since $W_{-1}(x) < -1$ for $x\in[-1/e^2,0)$, the denominator of the expression for $f'_{\gamma}(x)$ in Eq. \eqref{eqn:f_derive} is negative. To study the behavior of the numerator, we consider  another $\gamma$-parameterized function, namely $g_{\gamma}(x)$, as follows:
\begin{equation}
    g_{\gamma}(x)= \log(-x)-\gamma \left(1+W_{-1}(x) \right).
\label{eqn:g_defn}
\end{equation}
Using $g'_{\gamma}(x)$ to denote the derivative of $g_{\gamma}(x)$ w.r.t. $x$, we obtain
\begin{equation}
    g'_{\gamma}(x) = \frac{1+(1-\gamma)W_{-1}(x)}{x(1+W_{-1}(x))}.
    \label{eqn:g_deriv}
\end{equation}

\textbf{Establishing the lower bound:} For proving the lower bound on $W_{-1}(x)$ in \eqref{eqn:lambert_bnd}, we set $\gamma=4$ in Eq. \eqref{eqn:g_deriv} and note that both the numerator and the denominator of the resulting expression are positive; this once again follows from the fact that $W_{-1}(x) < -1$ for $x\in[-1/e^2,0)$. We conclude: $g'_4(x) > 0, \forall x \in [-1/e^2,0)$. Moreover, setting $\gamma=4$ in \eqref{eqn:g_defn}, we obtain
\begin{equation}
\begin{aligned}
g_4(-1/e^2) &= \log(1/e^2) - 4\left(1+W_{-1}(-1/e^2)\right) \\
&= -6 - 4 W_{-1}(-1/e^2) \\
& > -6 + 12 > 0,
\end{aligned}
\end{equation}
where we used $W_{-1}(-1/e^2) < -3$. We have thus shown that $g_4(x) > 0, \forall x\in [-1/e^2,0)$. Since the numerator of $f'_4(x)$ in Eq. \eqref{eqn:f_derive} is precisely $g_4(x)$, we conclude that $f'_4(x) < 0, \forall x\in [-1/e^2,0).$ At the same time, plugging in $\gamma=4$ in \eqref{eqn:f_defn} yields:
\begin{equation}
\begin{aligned}
f_4(-1/e^2) &= \log(1/e^2) - 4 \log(-\log(1/e^2))-W_{-1}(-1/e^2) \\
&= -2 - 4 \log(2) - W_{-1}(-1/e^2) \\
& < 2 - 4 \log(2) < 0,
\end{aligned}
\end{equation}
where we used $W_{-1}(-1/e^2) > -4$. Combining all the above pieces together, we conclude that $f_4(x) < 0, \forall x\in [-1/e^2,0)$. This immediately leads to the desired lower bound. 

\textbf{Establishing the upper bound:} The proof of the upper bound proceeds in a very similar way. We start by noting that
\begin{equation}
    g'_{1/4}(x) = \frac{1+3/4 W_{-1}(x)}{x(1+W_{-1}(x))}.
\end{equation}
For $x \in [-1/e^2,0)$, while the denominator of the fraction on the R.H.S. of the above expression is positive, the numerator is negative. The latter claim follows from the following facts: (i) $W_{-1}(-1/e^2) < -3$, and (ii) $W_{-1}(x)$ is monotonically decreasing. Thus, $g'_{1/4}(x) < 0, \forall x\in [-1/e^2,0).$ Moreover, 
\begin{equation}
\begin{aligned}
g_{1/4}(-1/e^2) &= \log(1/e^2) - \frac{1}{4}\left(1+W_{-1}(-1/e^2)\right) \\
&= - \frac{9}{4} - \frac{1}{4} W_{-1}(-1/e^2) \\
& < 0,
\end{aligned}
\end{equation}
where we used $W_{-1}(-1/e^2) > -4$. We have thus shown that $g_{1/4}(x) < 0, \forall x \in [-1/e^2,0)$. Since the numerator of $f'_{1/4}(x)$ is $g_{1/4}(x)$, it follows that $f'_{1/4}(x) > 0, \forall x\in [-1/e^2,0)$. Finally, observe
\begin{equation}
\begin{aligned}
f_{1/4}(-1/e^2) &= \log(1/e^2) - \frac{1}{4} \log(-\log(1/e^2))-W_{-1}(-1/e^2) \\
&= -2 - \frac{1}{4}\log(2) - W_{-1}(-1/e^2) \\
& > 1 - \frac{1}{4} \log(2) > 0,
\end{aligned}
\end{equation}
where we used $W_{-1}(-1/e^2) < -3$. We thus conclude that $f_{1/4}(x) > 0, \forall x\in [-1/e^2,0)$. The upper bound on $W_{-1}(x)$ in Eq. \eqref{eqn:lambert_bnd} follows immediately. This concludes the proof. 
\end{proof} 

\newpage
\section{Proof of Theorem \ref{thm:rFedSEL}} 
\label{app:RFedSELproof}
In this section, we analyze  \texttt{Robust Fed-SEL}, namely Algorithm \ref{algo:algo2}. Recall that $\mc{H}$ and $\mc{J}$ represent the sets of honest and adversarial clients, respectively. To proceed, let $T_{i,\ell}$ denote the total number of times an arm $\ell\in\mc{A}_j$ is pulled by an honest client $i\in\mc{C}_j \cap \mc{H}$ during the execution of \texttt{Robust Fed-SEL}. We then have the following analog of Lemma \ref{lemma:good_event}. 

\begin{lemma}
\label{lemma:Rgood_event}
 Consider the event $\mc{E_H}$ defined as follows:
\begin{equation}
    \mc{E_H} \triangleq \bigcap_{j\in [N]} \bigcap_{i\in \mc{C}_j \cap \mc{H}} \bigcap_{\ell \in\mc{A}_j} \bigcap_{t\in [T_{i,\ell}]} \{ \vert \hat{r}_{i,\ell}(t) - r_{\ell} \vert \leq \alpha_j(t) \}. 
\label{eqn:Revent}
\end{equation}
Then, it holds that $\mathbb{P}(\mc{E_H}) \geq 1-\delta$. 
\end{lemma}
\begin{proof}
Following the same reasoning as in Lemma \ref{lemma:good_event}, 
\begin{equation}
    \begin{aligned}
    \mathbb{P}(\mc{E}^c_{\mc{H}}) &\leq \sum_{j\in [N]} \sum_{i\in \mc{C}_j \cap \mc{H}} \sum_{\ell \in \mc{A}_j} \sum\limits_{t=1}^{\infty}  \mathbb{P} \left( \{\vert \hat{r}_{i,\ell}(t) - r_{\ell} \vert > \alpha_j(t) \} \right) \\
    & \leq \sum_{j\in [N]} \sum_{i\in \mc{C}_j \cap \mc{H}} \sum_{\ell \in \mc{A}_j} \sum\limits_{t=1}^{\infty} 2 \exp(-2 t \alpha^2_j(t)) \\
    & \leq \sum_{j\in [N]} \sum_{i\in \mc{C}_j \cap \mc{H}} \sum_{\ell \in \mc{A}_j} \sum\limits_{t=1}^{\infty} \frac{\delta}{2 |\mc{C}| |\mc{A}_j| t^2}\\
    & \leq \sum_{j\in [N]} \sum_{i\in \mc{C}_j \cap \mc{H}} \sum_{\ell \in \mc{A}_j} \frac{\delta}{2 |\mc{C}| |\mc{A}_j|} \left(1+\int_{1}^{\infty} \frac{1}{z^2} dz\right) \\
    & = \sum_{j\in [N]} \sum_{i\in \mc{C}_j \cap \mc{H}} \sum_{\ell \in \mc{A}_j} \frac{\delta}{|\mc{C}| |\mc{A}_j|} \\
    & =\frac{|\mc{C}\cap\mc{H}|}{|\mc{C}|}\delta \leq \delta,
    \end{aligned}
\end{equation}
which leads to the desired result. 
\end{proof}

The next lemma provides the main intermediate  result we need to prove Theorem \ref{thm:rFedSEL}.

\begin{lemma}
\label{lemma:robustmeans}
Suppose \texttt{Robust Fed-SEL} is run by every honest client. Moreover, suppose $|\mathcal{C}_j\cap \mathcal{J}| \leq f$ and $|\mathcal{C}_j| \geq 3f+1, \forall j\in\mathcal [N]$. On event $\mc{E_H}$, the following then hold: 
\begin{itemize}
    \item[(i)] $\bar{a}_j = j_1, \forall j\in [N]$.
    \item[(ii)] In very round $k$,
    \begin{equation}
       \bar{r}^{(U)}_j(k) \in \texttt{Conv}\left(\{\tilde{r}^{(U)}_{i}(k)\}_{i\in \bar{\mc{C}}_j \cap \mc{H}}\right); \hspace{2mm} \bar{r}^{(L)}_j(k) \in \texttt{Conv}\left(\{\tilde{r}^{(L)}_{i}(k)\}_{i\in \bar{\mc{C}}_j \cap \mc{H}}\right), \forall j\in[N],
\label{eqn:conv1}
    \end{equation}
\end{itemize}
where we use $\texttt{Conv}(\Omega)$ to represent the convex hull of a set of points $\Omega$. 
\end{lemma}
\begin{proof}
Fix a client group $j\in[N].$ Based on Lemma \ref{lemma:Algo1}, note that when each client $i\in\mc{C}_j \cap \mc{H}$ runs Algo. \ref{algo:algo1} on $\mc{A}_j$, the local subroutine will eventually terminate on event $\mc{E_H}$, and client $i$ will output $a_i = j_1$, i.e., each honest client $i$ in $\mc{C}_j$ will correctly identify the locally best arm $j_1$ in $\mc{A}_j$ at the end of Phase I. Since $|\mathcal{C}_j\cap \mathcal{J}| \leq f$ and $|\mathcal{C}_j| \geq 3f+1$, at least $2f+1$ clients from $\mc{C}_j$ are guaranteed to report to the server at the end of Phase I. Among them, since at most $f$ can be adversarial, observe that the  majority voting operation in line 2 of \texttt{Robust Fed-SEL} yields $\bar{a}_j = j_1$. This establishes the first claim of the lemma.

For the second claim, let us start by noting that on $\mc{E_H}$, the number of adversaries in the set $\bar{\mc{C}}_j$ is at most $f_j = f-|\mc{C}'_j| + |\bar{\mc{C}}_j|$. This follows from the fact that on $\mc{E_H}$, any client $i\in \mc{C}'_j$ who reports an arm other than $j_1$ must be adversarial, i.e., on $\mc{E_H}$, $\mc{C}'_j \setminus \bar{\mc{C}}_j \subseteq \mc{J}$. Moreover, observe that
\begin{equation}
    \begin{aligned}
    2 f_j+1 &= 2 \left(f-|\mc{C}'_j| + |\bar{\mc{C}}_j|\right)+1\\
    & = 2\left(f -(2f+1) + |\bar{\mc{C}}_j|\right)+1\\
    & = \left(|\bar{\mc{C}}_j|-(2f+1) \right) + |\bar{\mc{C}}_j| \\
    & \leq  \left(|\mc{C}'_j|-(2f+1) \right) + |\bar{\mc{C}}_j| \\
    & = |\bar{\mc{C}}_j|. 
    \end{aligned}
\end{equation}

With these basic observations in place, we are now ready to analyze the \texttt{Trim} operation outlined in line 6 of \texttt{Robust Fed-SEL}. To establish the claim that 
$$
    \bar{r}^{(U)}_j(k) \in \texttt{Conv}\left(\{\tilde{r}^{(U)}_{i}(k)\}_{i\in \bar{\mc{C}}_j \cap \mc{H}}\right),
$$
 suppose $\texttt{Trim}\left( \{\tilde{r}^{(U)}_{i}(k)\}_{i\in\bar{\mc{C}}_j},f_j \right) = \Tilde{r}^{(U)}_{w}(k)$, where $w \in \bar{\mc{C}}_j$. If $w\in\mc{H}$, then the claim holds  trivially. Now suppose $w$ is adversarial, i.e., $w\in \mc{J}$. Since $|\bar{\mc{C}}_j \cap \mc{J} | \leq f_j$, and $|\bar{\mc{C}}_j| \geq 2 f_j +1$, there must exist honest clients $u,v \in \bar{\mc{C}}_j \cap \mc{H}$, such that $\Tilde{r}^{(U)}_{v}(k) \leq \Tilde{r}^{(U)}_{w}(k) \leq \Tilde{r}^{(U)}_{u}(k)$. This establishes the claim about $\bar{r}^{(U)}_j(k)$ in Eq. \eqref{eqn:conv1}; the claim regarding $\bar{r}^{(L)}_j(k)$ follows the exact same argument.
\end{proof}

The above lemma tells us that with high probability, (i) the server will correctly identify the locally best arm from each arm-set, and (ii) the thresholds used for eliminating client-groups are ``not too corrupted" by the adversaries. We now show how these two properties feature in the proof of Theorem \ref{thm:rFedSEL}.

\begin{proof} (\textbf{Theorem \ref{thm:rFedSEL}}) 
Throughout the proof, we will condition on the event $\mc{E_H}$ defined in Eq. \eqref{eqn:Revent}. Based on Lemma \ref{lemma:Rgood_event}, recall that $\mc{E_H}$  occurs with probability at least $1-\delta$. As in the proof of Theorem \ref{thm:Fed-SEL}, we start by arguing that client group $1$ always remains active (recall that $a^*=1 \in \mc{A}_1$). To see why this is true, fix any $j\in [N]$, and observe that for each $i\in\bar{\mc{C}}_j \cap \mc{H}$ and every round $k$, 
\begin{equation}
\begin{aligned}
    \tilde{r}^{(L)}_{i}(k) &=\tilde{r}_i(k) - 2 \alpha_j(\bar{t}_i+kH) \\ 
    & \overset{(a)} \leq \hat{r}_{i,\bar{a}_j}(\bar{t}_i +kH) + \frac{\alpha_j(\bar{t}_i+kH)}{2} - 2 \alpha_j(\bar{t}_i+kH) \\
    & \overset{(b)}= \hat{r}_{i,j_1}(\bar{t}_i+kH) - \frac{3}{2}\alpha_j(\bar{t}_i+kH)\\
    &\overset{\mc{E}_\mc{H}}\leq {r_{j_1}} + \alpha_j(\bar{t}_i+kH)-\frac{3}{2}\alpha_j(\bar{t}_i+kH)\\
    & < r_{j_1}.
\label{eqn:lowbnd}
\end{aligned}
\end{equation}
In the above steps, (a) follows from (i) the fact that by definition of the set $\bar{\mc{C}}_j$, every client $i$ in $\bar{\mc{C}}_j$ reports $a_i=\bar{a}_j$, and (ii) by using the bound on the quantization error in Eq. \eqref{eqn:quant_err}. Based on Lemma \ref{lemma:robustmeans}, the majority voted arm $\bar{a}_j$ is $j_1$ on event $\mc{E_H}$; this leads to $(b)$. Using the same reasoning, we can show that for each $i\in\bar{\mc{C}}_j \cap \mc{H}$ and every round $k$, 
\begin{equation}
\tilde{r}^{(U)}_{i}(k) > r_{j_1}
\label{eqn:uppbnd}
\end{equation}
on event $\mc{E_H}$. Appealing to  \eqref{eqn:conv1} from Lemma \ref{lemma:robustmeans}, and using  \eqref{eqn:lowbnd} and \eqref{eqn:uppbnd}, we conclude that on $\mc{E_H}$, 
$$ \bar{r}^{(U)}_1(k) > r_1; \hspace{4mm} \bar{r}^{(L)}_j(k) < r_{j_1}, \forall j\in [N]\setminus \{1\}.$$
From lines 7 and 8 of \texttt{Robust Fed-SEL}, we immediately note that client-group $1$ never gets eliminated. To show that \texttt{Robust Fed-SEL} is consistent, it remains to argue that on $\mc{E_H}$, every client-group $j\in [N]\setminus \{1\}$ eventually gets eliminated. We claim that for group $j\in [N]\setminus \{1\}$ to get eliminated, the following is a set of sufficient conditions:
\begin{equation}
    \begin{aligned}
    \alpha_j(\bar{t}_i+kH) &\leq \frac{\Delta_{1j_1}}{8}, \forall i \in  \bar{\mc{C}}_j \cap \mc{H}, \textrm{and} \\
    \alpha_1(\bar{t}_i+kH) &\leq \frac{\Delta_{1j_1}}{8}, \forall i \in  \bar{\mc{C}}_1 \cap \mc{H}.
    \end{aligned}
\label{eqn:Rsuff_cond}
\end{equation}
Suppose the above conditions hold. Then, based on the same reasoning we used to arrive at Eq. \eqref{eqn:FSEL_ub2}, we can show that for each $i \in  \bar{\mc{C}}_j \cap \mc{H}$, 
\begin{equation}
\begin{aligned}
    \tilde{r}^{(U)}_{i}(k) \leq r_{j_1}+\frac{7}{16} \Delta_{1j_1}, 
\end{aligned}
\label{eqn:uppbnd2}
\end{equation}
and for each $i \in  \bar{\mc{C}}_1 \cap \mc{H}$, 
\begin{equation}
    \tilde{r}^{(L)}_{i}(k) \geq r_{1} - \frac{7}{16}\Delta_{1j_1}.
\label{eqn:lowbnd2}
\end{equation}
Appealing once again to the second claim of Lemma \ref{lemma:robustmeans}, and using \eqref{eqn:uppbnd2} and \eqref{eqn:lowbnd2}, we conclude that if the conditions in \eqref{eqn:Rsuff_cond} hold, then
\begin{equation}
    \bar{r}^{(L)}_1(k) \geq r_1 - \frac{7}{16}{\Delta_{1j_1}}\hspace{1mm} \textrm{and} \hspace{1mm} \bar{r}^{(U)}_j(k) \leq r_{j_1} +  \frac{7}{16}{\Delta_{1j_1}},
\end{equation}
implying $\bar{r}^{(L)}_1(k) > \bar{r}^{(U)}_j(k)$. From line 8 of \texttt{Robust Fed-SEL}, this is sufficient for client-group $j$ to get eliminated. Thus, \texttt{Robust Fed-SEL} will eventually terminate on $\mc{E_H}$ with $\bar{\Psi}(k+1)=\{1\}$, for some round $k$. Consistency then follows from Lemma \ref{lemma:robustmeans} by noting that on $\mc{E_H}$, $\bar{a}_1=a^*=1$.

The sample- and communication-complexity bounds of \texttt{Robust Fed-SEL} are of the same order as that of \texttt{Fed-SEL}, and can be derived in exactly the same way as in Theorem \ref{thm:Fed-SEL}. We now briefly explain how this can be done. Suppose we want to compute an upper bound on the number of rounds client group $j\in[N]\setminus\{1\}$ remains active. Start by noting that at the end of Phase I, the following bound applies \textit{uniformly} to every  $i\in\bar{\mc{C}}_j\cap\mc{H}$:
$$ \alpha_j(\bar{t}_i) < \frac{2}{9} \Delta^*_j <   \frac{\sigma_{j1} \Delta_{1j_1}}{8}.$$
This follows from Lemma \ref{lemma:alpha_bnd}.
 For each $i\in\bar{\mc{C}}_j\cap\mc{H}$, let $R^{(j)}_{i1}$ be the first round such that
$$ \alpha_j(\bar{t}_i+kH) \leq \frac{\Delta_{1j_1}}{8}.$$
We can obtain an upper bound on $R^{(j)}_{i1}$ for each $i\in\bar{\mc{C}}_j\cap\mc{H}$ exactly as in Theorem \ref{thm:Fed-SEL}, and show that
$$ \max_{i\in \bar{\mc{C}}_j\cap\mc{H}} R^{(j)}_{i1} \leq \frac{1}{H}\left( 
    \frac{128}{{(\Delta^*_j)}^2} (\sigma^2_{j1} -1) \log{\frac{128\bar{c}_j}{\Delta^2_{1j_1}}} + \frac{256}{{(\Delta^*_j)}^2} \log{\sigma_{j1}} + O\left(\frac{1}{\Delta^2_{1j_1}}\log\log{\frac{\bar{c}_j}{\Delta^2_{1j_1}}} \right) \right)+1, \hspace{1mm} \textrm{if} \hspace{1mm} \sigma_{j1} > 1, $$
and $\max_{i\in \bar{\mc{C}}_j\cap\mc{H}} R^{(j)}_{i1} =1$, if $\sigma_{j1} \leq 1.$ With $R_{j1} \triangleq \max_{i\in \bar{\mc{C}}_j\cap\mc{H}} R^{(j)}_{i1}$, observe that the expression for $R_{j1}$ exactly resembles that for $R_{i1}$ in Eq. \eqref{eqn:Ri1} of Theorem \ref{thm:Fed-SEL}.\footnote{Recall that in the context of Theorem \ref{thm:Fed-SEL}, each client group contained only one agent, and hence the index $i$ was used for both client groups and agents.} Proceeding as we did in Theorem \ref{thm:Fed-SEL}, we can now derive an upper bound on the number of rounds group $j$ remains active. The rest of the proof mirrors that of Theorem \ref{thm:Fed-SEL}, and hence we omit details. 
\end{proof}

\section{Proof of Theorem \ref{thm:peertopeer}}
In this section, we will analyze the peer-to-peer setting described in Section \ref{sec:MAB_network}. As before, we need a ``good event" on which we will base our reasoning. For the setting we consider here, the event $\mc{E_H}$ described in Eq. \eqref{eqn:Revent} will serve such a purpose. Recall that $\mathbb{P}(\mc{E_H}) \geq 1-\delta.$ 

Consider any honest client $i\in \mc{C}_j.$ The argument we present next applies identically to honest clients in other client groups as well. Since $i\in\mc{C}_j$, running Algo. \ref{algo:algo1} on $\mc{A}_j$ with $\beta_j(t)=6\alpha_j(t)$ yields

\begin{equation}
    a_i = j_1; \hspace{1mm} \vert \hat{r}_{i,a_i}(\bar{t}_i) - r_{j_1} \vert =  \vert \hat{r}_{i,j_1}(\bar{t}_i) - r_{j_1} \vert \leq \alpha_j(\bar{t}_i) < \frac{c}{(c-2)^2} \Delta^*_j= \frac{3}{8} \Delta^*_j
    \label{eqn:p2p1}
\end{equation}
on event $\mc{E_H}$. The above claims follow directly from Lemma's \ref{lemma:Algo1} and \ref{lemma:alpha_bnd}. Thus, based on Line 1 of Algo. \ref{algo:p2p}, we have $\mb{I}_i(j) = j_1$, and $\mb{R}_i(j) = \hat{r}_{i,j_1}(\bar{t}_i)$, where $\hat{r}_{i,j_1}(\bar{t}_i)$ satisfies the bound in Eq. \eqref{eqn:p2p1}. 

Now for client $i$ to identify the globally best arm, it must eventually receive accurate information about every other arm-set $\ell \in [N] \setminus \{j\}$ from its neighbors, despite the presence of adversaries in the network. The next lemma shows how the strong-robustness property in Definition \ref{def:strongrobustness} contributes to this cause. 

\begin{lemma} 
\label{lemma:percolation}
Suppose Algo. \ref{algo:p2p} is run by every $i\in\mc{H}$. Moreover, suppose the adversary model is $f$-local, and $\mc{G}$ is strongly-robust w.r.t. $\mc{C}_j, \forall j\in [N]$. Consider any $j\in [N]$. Then, on the event $\mc{E_H}$, every client $i\in \mc{V}\setminus\mc{C}_j$ receives information about arm-set $j$ from at least $2f+1$ neighbors in $\mc{N}_i$. 
\end{lemma}
\begin{proof}
We prove this lemma by contradiction. Fix any $j\in [N]$. Suppose that on the event $\mc{E_H}$, there exists a non-empty set $\mc{B} \subseteq \mc{V} \setminus \mc{C}_j$ of clients who never receive information about arm-set $j$ from $2f+1$ neighbors. Based on the strong-robustness property in Def. \ref{def:strongrobustness}, there exists a client $i\in \mc{B}$ with at least $3f+1$ neighbors in $\mc{V} \setminus \mc{B}$. Consider the set $\mc{Q}_{i,j} = \mc{N}_i \cap \{\mc{V} \setminus \mc{B}\}$. Based on the strong-robustness property and the $f$-local assumption, $|\mc{Q}_{i,j}\cap \mc{H}| \geq 2f+1$. Now a client $k$ in $\mc{Q}_{i,j}\cap \mc{H}$ can either belong to $\mc{C}_{j}$ or not. If $k \in \mc{Q}_{i,j}\cap \mc{H}$ belongs to $\mc{C}_j$, then on the event $\mc{E_H}$,  Algo. \ref{algo:algo1} will terminate at client $k$. Thus, client $k$ will eventually transmit $\{\mb{I}_k(j), \mb{R}_k(j)\}$ to every out-neighbor in  $\mc{N}^{+}_k$, including client $i$. Even if $k \in \mc{Q}_{i,j}\cap \mc{H}$ does not belong to $\mc{C}_j$, since $k \notin \mc{B}$, client $k$ must have received information about arm-set $j$ from $2f+1$ in-neighbors. Thus, based on lines 2-4 of Algo. \ref{algo:p2p}, it must transmit $\{\mb{I}_k(j), \mb{R}_k(j)\}$ to client $i$ at some point of  time. We conclude that on the event $\mc{E_H}$, client $i$ will eventually receive information about arm-set $j$ from every neighbor in  $\mc{Q}_{i,j}\cap \mc{H}$. This leads to the desired contradiction and concludes the proof. 
\end{proof}

 Our next goal is to prove that on $\mc{E_H}$, for each $j \in [N]$, every honest client $i\in \mc{V}\setminus\mc{C}_j$ is able to correctly identify the best arm $j_1$ in $\mc{A}_j$, and obtain a reasonable estimate of its mean $r_{j_1}$. We establish these properties in the next lemma. 
 
 \begin{lemma}
 \label{lemma:robust_estimates}
 Suppose Algo. \ref{algo:p2p} is run by every $i\in\mc{H}$. Moreover, suppose the adversary model is $f$-local, and $\mc{G}$ is strongly-robust w.r.t. $\mc{C}_j, \forall j\in [N]$. On event $\mc{E_H}$, the following then hold $\forall j\in [N]$: 
\begin{itemize}
    \item[(i)] $\mb{I}_i(j) = j_1, \forall i\in\mc{H}$. 
    \item[(ii)] $\vert \mb{R}_i(j) - r_{j_1} \vert \leq (3/8)  \Delta^*_j, \forall i \in \mc{H}.$
\end{itemize}
 \end{lemma}
 \begin{proof}
 Fix any $j\in [N]$. If $i \in \mc{C}_j \cap \mc{H}$, then each of the above claims hold based on the arguments we presented at the beginning of this section; see Eq. \eqref{eqn:p2p1}. Thus, we only need to prove these claims for clients in $\{\mc{V}\setminus\mc{C}_j\} \cap \mc{H}$. 
 
 To do so, let us first associate a notion of ``activation" with Algo. \ref{algo:p2p}. For each client $i\in \mc{C}_j \cap \mc{H}$, we say that it gets activated when the local subroutine (Algo. \ref{algo:algo1}) terminates at client $i$. For each $i \in \{\mc{V}\setminus\mc{C}_j\} \cap \mc{H}$, we say it gets activated upon receiving information about arm-set $j$ from $2f+1$ neighbors in $\mc{N}_i$. From Lemma's \ref{lemma:Algo1} and \ref{lemma:percolation}, observe that all honest clients eventually get activated on the event $\mc{E_H}$. Accordingly, on the event $\mc{E_H}$, let $\tau_1 < \tau_2 < \cdots < \tau_K$ denote the activation times for clients in $\{\mc{V}\setminus\mc{C}_j\} \cap \mc{H}$.\footnote{Note that multiple clients may get activated at the same time-step.} Moreover, let $\mc{H}^{(j)}_{\tau_s}$ represent those clients in $\{\mc{V}\setminus\mc{C}_j\} \cap \mc{H}$ that get activated at time-step $\tau_s, s \in [K]$.
 
 We prove the result by inducting on the activation times. For the base case, consider any client $i$ in the set $\mc{H}^{(j)}_{\tau_1}$. Recall that $\mc{N}'_{i,j} \subseteq \mc{N}_i$ are the first $2f+1$ neighbors who report about arm-set $j$ to client $i$. From the rules of Algo. \ref{algo:p2p}, notice that an honest client $k$ transmits $\{\mb{I}_k(j), \mb{R}_k(j)\}$ only when it gets activated. Thus, it must be that $\mc{N}'_{i,j} \cap \mc{H} \subseteq \mc{C}_j \cap \mc{H}$. Moreover, since the adversary model is $f$-local, $|\mc{N}'_{i,j} \cap \mc{H}| \geq f+1$. We conclude that at least $f+1$ honest clients in $\mc{C}_j$ report arm $j_1$ to client $i$. Based on the majority voting operation in line 3 of Algo. \ref{algo:p2p}, it follows that $\mb{I}_i(j)=j_1$. 
 
 Observe also that on the event $\mc{E_H}$, any client $k \in \mc{N}'_{i,j}$ that reports an arm other than arm $j_1$ to client $i$ must be adversarial. Following the same reasoning as in Lemma \ref{lemma:robustmeans}, we can then establish that (i) $|\bar{\mc{N}}_{i,j}| \geq 2 f_{i,j} +1$, where $f_{i,j} = f-|\mc{N}'_{i,j}| + |\bar{\mc{N}}_{i,j}|$, and $\bar{\mc{N}}_{i,j}$ represents those clients in $\mc{N}'_{i,j}$ that report arm $j_1$; and (ii) $|\bar{\mc{N}}_{i,j} \cap \mc{J}| \leq f_{i,j}$. Using the same arguments used to arrive at Eq. \eqref{eqn:conv1}, we can then show that
 \begin{equation}
   \mb{R}_i(j) \in \texttt{Conv}\left(\{\mb{R}_k(j)\}_{k\in \bar{\mc{N}}_{i,j} \cap \mc{H}}\right) \implies  \mb{R}_i(j) \in \texttt{Conv}\left(\{\mb{R}_k(j)\}_{k\in {\mc{C}}_j \cap \mc{H}}\right),
   \label{eqn:conv2}
 \end{equation}
 since $\bar{\mc{N}}_{i,j} \cap \mc{H} \subseteq {\mc{C}}_j \cap \mc{H}$. For client $i$, claim (ii) of the lemma then follows immediately from the fact that on event $\mc{E_H}$, $\vert \mb{R}_k(j) - r_{j_1} \vert \leq (3/8)  \Delta^*_j, \forall k \in {\mc{C}}_j \cap \mc{H}$. This completes the base case.
 
 For the induction hypothesis, suppose for all $i\in \bigcup_{s\in[m]} \mc{H}^{(j)}_{\tau_s}$, $m\in [K-1]$,  it holds that $\mb{I}_i(j)=j_1$, and 
 \begin{equation}
 \mb{R}_i(j) \in \texttt{Conv}\left(\{\mb{R}_k(j)\}_{k\in {\mc{C}}_j \cap \mc{H}}\right).
 \label{eqn:conv3}
 \end{equation}
 
 Now consider a client $i\in \mc{H}^{(j)}_{\tau_{m+1}}$. Based on the rules of Algo. \ref{algo:p2p}, and the definition of activation times, it must be that
 $$ \mc{N}'_{i,j} \cap \mc{H} \subseteq \{\mc{C}_j \cap \mc{H}\} \cup \{\bigcup_{s\in[m]} \mc{H}^{(j)}_{\tau_s}\}. $$
 Using the above fact, $|\mc{N}'_{i,j} \cap \mc{H}| \geq f+1$, the induction hypothesis, and the same reasoning as the base case, it is easy to see that $\mb{I}_i(j)=j_1$, and that $\mb{R}_i(j)$ satisfies the inclusion in Eq. \eqref{eqn:conv3}. This completes the induction hypothesis and the proof. 
 \end{proof}
 
 We are now ready to complete the proof of Theorem \ref{thm:peertopeer}.
 
 \begin{proof} \textbf{(Theorem \ref{thm:peertopeer})}  To complete the proof of Theorem \ref{thm:peertopeer}, it remains to argue that for each honest client $i$, $\mb{R}_i(1) > \mb{R}_i(j), \forall j \in [N]\setminus \{1\}$. By way of contradiction, suppose there exists some client $i\in\mc{H}$ and some $j\in [N] \setminus \{1\}$, such that $\mb{R}_i(j) \geq \mb{R}_i(1)$. We then have
\begin{equation}
    \begin{aligned}
     & \mb{R}_i(j) \geq \mb{R}_i(1) \\
\overset{(a)}\implies & r_{j_1} + \frac{3}{8} \Delta^*_j \geq r_1 - \frac{3}{8} \Delta^*_1 \\
\implies & \Delta_{1j_1} \leq  \frac{3}{8} (\Delta^*_1+\Delta^*_j) \\
\overset{(b)}\implies & \Delta_{1j_1} \leq  \frac{3}{8} \left(\sigma_{1j}+\sigma_{j1}\right)\Delta_{1j_1} \\
\overset{(c)}\implies & \Delta_{1j_1} \leq \frac{3}{4} \Delta_{1j_1},
    \end{aligned}
\end{equation}
which leads to the desired contradiction. Here, (a) follows from the second claim of Lemma \ref{lemma:robust_estimates}, (b) follows from the definition of arm-heterogeneity indices in Eq. \eqref{eqn:arm_het}, and (c) follows from the assumption on arm-heterogeneity in Eq. \eqref{eqn:het_ass}. This completes the proof. 
  \end{proof}
 
\end{document}